\documentclass[runningheads]{llncs}
\usepackage{graphicx}

\usepackage{tikz}
\usepackage{comment}
\usepackage{amsmath,amssymb} 
\usepackage{color}
\usepackage[accsupp]{axessibility}  

\usepackage[misc]{ifsym}
\usepackage[toc,page]{appendix}

\usepackage{booktabs}
\usepackage{bm}
\usepackage{soul}
\usepackage{cuted}
\usepackage{xcolor}
\usepackage{caption}
\usepackage{multirow}

\usepackage{pifont}
\newcommand{\xmark}{\ding{55}}

\usepackage{array}
\newcommand{\PreserveBackslash}[1]{\let\temp=\\#1\let\\=\temp}
\newcolumntype{C}[1]{>{\PreserveBackslash\centering}p{#1}}
\newcolumntype{R}[1]{>{\PreserveBackslash\raggedleft}p{#1}}
\newcolumntype{L}[1]{>{\PreserveBackslash\raggedright}p{#1}}

\begin{document}
\pagestyle{headings}
\mainmatter

\title{Prompting Visual-Language Models for Efficient Video Understanding}

\titlerunning{Prompt Learning for Efficient Video Understanding}

\author{Chen Ju\inst{1} \and
Tengda Han\inst{2} \and
Kunhao Zheng\inst{1} \and
Ya Zhang\inst{1} \and
Weidi Xie\inst{1,2}}

\authorrunning{C. Ju et al.}

\institute{Cooperative Medianet Innovation Center, Shanghai Jiao Tong University \and
Visual Geometry Group, University of Oxford \\
\email{ \{ju\_chen,\,dyekuu,\,ya\_zhang,\,weidi\}@sjtu.edu.cn \ htd@robots.ox.ac.uk} \\
\url{https://ju-chen.github.io/efficient-prompt}}

%******************
\maketitle

\begin{abstract}
Image-based visual-language (I-VL) pre-training has shown great success for learning joint visual-textual representations from large-scale web data, revealing remarkable ability for ``zero-shot'' generalisation. This paper presents a simple but strong baseline to efficiently adapt the pre-trained I-VL model, and exploit its powerful ability for resource-hungry video understanding tasks, with minimal training. Specifically, we propose to optimise a few random vectors, termed as ``continuous prompt vectors'', that convert video-related tasks into the same format as the pre-training objectives. In addition, to bridge the gap between static images and videos, temporal information is encoded with lightweight Transformers stacking on top of frame-wise visual features. Experimentally, we conduct extensive ablation studies to analyse the critical components. On $10$ public benchmarks of action recognition, action localisation, and text-video retrieval, across closed-set, few-shot, and zero-shot scenarios, we achieve competitive or state-of-the-art performance to existing methods, despite optimising significantly fewer parameters.
\end{abstract}

\section{Introduction}  \label{sec:intro}
While the research in computer vision has mainly focused on tackling particular tasks, the grand goal towards human-level perception has always been to learn general-purpose visual representation, that can solve various problems with {\em minimal tunings}. Towards such a goal, recent work for training image-based visual-language \textbf{(I-VL)} models has shown promising progress. For example, CLIP~\cite{Radford21} and ALIGN~\cite{Jia21} learn the joint representation for image and text with simple noise contrastive learning, greatly benefiting from the rich information in text descriptions, {\em e.g.} actions, objects, human-object interactions, and object-object relationships. As a result, these pre-trained I-VL models have demonstrated remarkable ``zero-shot'' generalisation for various image classification tasks. Crucially, the data used to train these powerful I-VL models can simply be crawled from the Internet at scale, without any laborious manual annotation. It is therefore reasonable to believe, with the growing computation, larger datasets will be collected, and more powerful models will be trained in the near future.

Given this promise, one question naturally arises: {\em how can we best exploit the ability in the powerful I-VL models, and effectively adapt it to solve novel vision tasks of interest?} One possible solution would be to finetune the image encoder end-to-end on the downstream tasks, however, since each task needs to finetune and save its own set of parameters, we end up developing hundreds of models for hundreds of individual tasks. Even more problematic, discarding the text encoder loses the model's ability for ``zero-shot'' generalisation, thus the resultant model can only work for a fixed set of pre-determined categories. Alternatively, as shown in CLIP~\cite{Radford21}, given properly designed ``prompts'', the model is able to work on a variety of downstream tasks including ``zero-shot'', with classifiers being dynamically generated by the text encoder, from category names or other free-form texts. The prompts here are handcrafted cloze templates to facilitate classifier generation, so that novel tasks can be formulated in the same format as pre-training objectives, effectively closing the gap between pre-training and downstream tasks. One remaining issue is, such handcrafted prompts require extensive expert knowledge and labor, limiting the use for efficient adaptation.

In this paper, we continue the vein of prompt-based learning~\cite{Lester21,li21-prefixtuning}, with the goal of exploring a comprehensive and strong baseline to adapt I-VL models for {\em efficient} video understanding. We here focus on the resource-hungry video tasks, for three reasons: 1) From the \textit{data} perspective, comparing to image-text pairs, video-text pairs are harder to collect, and may suffer from misalignment issues~\cite{han2022align};
\hspace{1pt} 2) Solving video tasks demands more computational power. Given the same budget, training on image-text pairs enables the model to learn more \textit{diversity}, making it more cost-effective to understand video with I-VL models;
\hspace{1pt} 3) Videos are composed of frame sequences, establishing temporal dependencies on powerful image-based models is a natural choice.

Specifically, we consider a simple idea by prepending\,/\,
appending a sequence of random vectors, termed as ``continuous prompt vectors'', 
to the textual input. These prompt vectors consist entirely of free parameters that do not correspond to any real concrete words, 
and the subsequent layers of the text encoder will attend these vectors, 
as if they were a sequence of ``virtual tokens'' to generate the corresponding classifier or embedding. During training, we freeze the weights of the I-VL text encoder, 
and the gradients are back-propagated to optimise these learnable prompt vectors. Consequently, a single copy of the visual backbone is able to perform various video tasks, with the minimal number of trainable parameters for each task. To further exploit the video temporal information, we also append {\em lightweight} Transformers on top of frame-wise visual representation. As a result, the various video tasks can be formulated under the same umbrella, {\em i.e.} to maximise the similarity matching between visual and textual embeddings, with texts being action category names or fine-grained descriptions.

To summarise, building on scalable and powerful I-VL models, we first propose a simple baseline for {\em efficient} and {\em lightweight} video understanding, by learning the task-specific prompt vectors, which facilitates possible future research in video action recognition, action localisation, and text-video retrieval; We extensively evaluate the efficient adaptation idea on ten public benchmarks, across closed-set, few-shot, and zero-shot scenarios, and thoroughly dissect the critical components;
Lastly, despite training only a few free parameters, {\em i.e.}~several prompt vectors and two Transformer layers, in the closed-set scenario, we achieve competitive or state-of-the-art performance to existing methods. In few-shot and zero-shot scenarios, we significantly outperform all previous methods on $7$ popular benchmarks, sometimes by over 10\% gains.

\section{Related Work} \label{sec:related}
{\noindent \bf Joint Visual-Textual Learning.} 
In the literature, \cite{Mori99} has explored the connection between images and words using paired text documents, and~\cite{Frome13,Weston11} proposed to jointly learn image-text embeddings with the category name annotations. Recently, CLIP~\cite{Radford21}, ALIGN~\cite{Jia21} and FILIP~\cite{yao2021filip} have further scaled up the training with large-scale web data. Using simple noise contrastive learning, it is shown that powerful visual representation can be learnt from paired image-caption. In video domains, similar idea has also been explored for representation learning~\cite{Miech20} and video retrieval~\cite{anne2017localizing,Lei21,Miech18}. In this paper, we establish baselines on steering the pre-trained CLIP model to video understanding tasks, the same technique should be applicable to other I-VL models as well.

\vspace{0.15cm}
{\noindent \bf Prompting}
refers to designing proper ``instructions'' that the pre-trained language model can understand, and generate desired outputs, using a few examples as demonstrations. For instance, given properly handcrafted prompt templates, GPT-3~\cite{Brown20} has shown strong generalisations for few-shot or zero-shot learning. However, the handcrafted templates require extensive expert knowledge, limiting the flexibility. Later work proposes to automate prompt engineering by searching discrete prompts~\cite{Gao21,Jiang20,Timo21,Shin20}, and continuous prompts~\cite{Lester21,li21-prefixtuning}. In this work, we consider to search continuous prompts for steering pre-trained visual-language models to tackle video understanding tasks.

\vspace{0.15cm}
{\noindent \bf Video Action Recognition.} 
Effective architecture research has gone through rapid developments, from the two-stream network~\cite{Feichtenhofer16,Simonyan14,Wang16-TSN} to more recent single stream RGB networks~\cite{Bertasius21,bulat2021space,carreira2017quo,Feichtenhofer20,Feichtenhofer19,Tran18,Xie18-S3D}. With the help of abundant data, {\em e.g.}~Kinetics~\cite{Carreira19}, recognition accuracy has been steadily improved.
In addition, data-efficient learning has also been explored: few-shot and zero-shot action recognition. Specifically, in the former research line, only a few training samples are available from each action category, \cite{Zhu18,zhu2020label,zhu2021few} proposed compound memory networks to classify videos by matching and ranking; \cite{Dwivedi19} used GANs to synthesize training examples for novel categories; \cite{Cao20} proposed differentiable dynamic time warping to align videos of different lengths; \cite{Perrett21} exploited a CrossTransformer to find temporally-corresponding frame tuples. While for zero-shot action recognition, it requires to generalise towards action categories that are unseen in the training set, one typical idea lies in learning a common representation space that is shared by seen and unseen categories, such as attributes space~\cite{Gan16a,jain201515,Liu11}, semantic space~\cite{brattoli2020rethinking,Gan16b,jain2015objects2action,Li16}, synthesizing features to unseen actions~\cite{Mishra20}, using objects to create common space for unseen actions~\cite{Mettes21}.

\vspace{0.15cm}
{\noindent \bf Video Action Localisation} aims to detect and classify actions in untrimmed long videos. In general, there are two popular detection paradigms: the two-stage paradigm~\cite{Chao18,ju2021divide,lin2019bmn,Lin18,Shou16,tan2021relaxed,wang2022rcl,Xu17,Zhao17} first localises class-agnostic action proposals, which covers correct segments with high recall, then classifies and refines each proposal. The one-stage paradigm~\cite{buch2019end,ju2021adaptive,lin2017single,Nawhal21,Yeung16,zhang2022actionformer} combines localisation and classification, {\em i.e.} densely classifies each frame into actions or background.

\vspace{0.15cm}
{\noindent \bf Concurrent Work.}
Several recent papers~\cite{gao2021clip,jia2022visual,zhang2021tip,zhou2021learning,zhou2022conditional} also explore prompt learning for efficient transfer from pre-trained CLIP to downstream image tasks. In the video domains,~\cite{Luo_CLIP4Clip21,Wang_ActionCLIP21} propose to end-to-end finetune CLIP on individual video tasks, {\em e.g.}~action recognition and retrieval. In contrast, we favor efficient adaptation from image to video, present the first yet simple approach on prompt learning, to establish strong and wide baselines for video understanding.

% ------------------------------------------------------
% ------------------------------------------------------
\begin{figure*}[t]
\begin{center}
\includegraphics[width=1.0\textwidth] {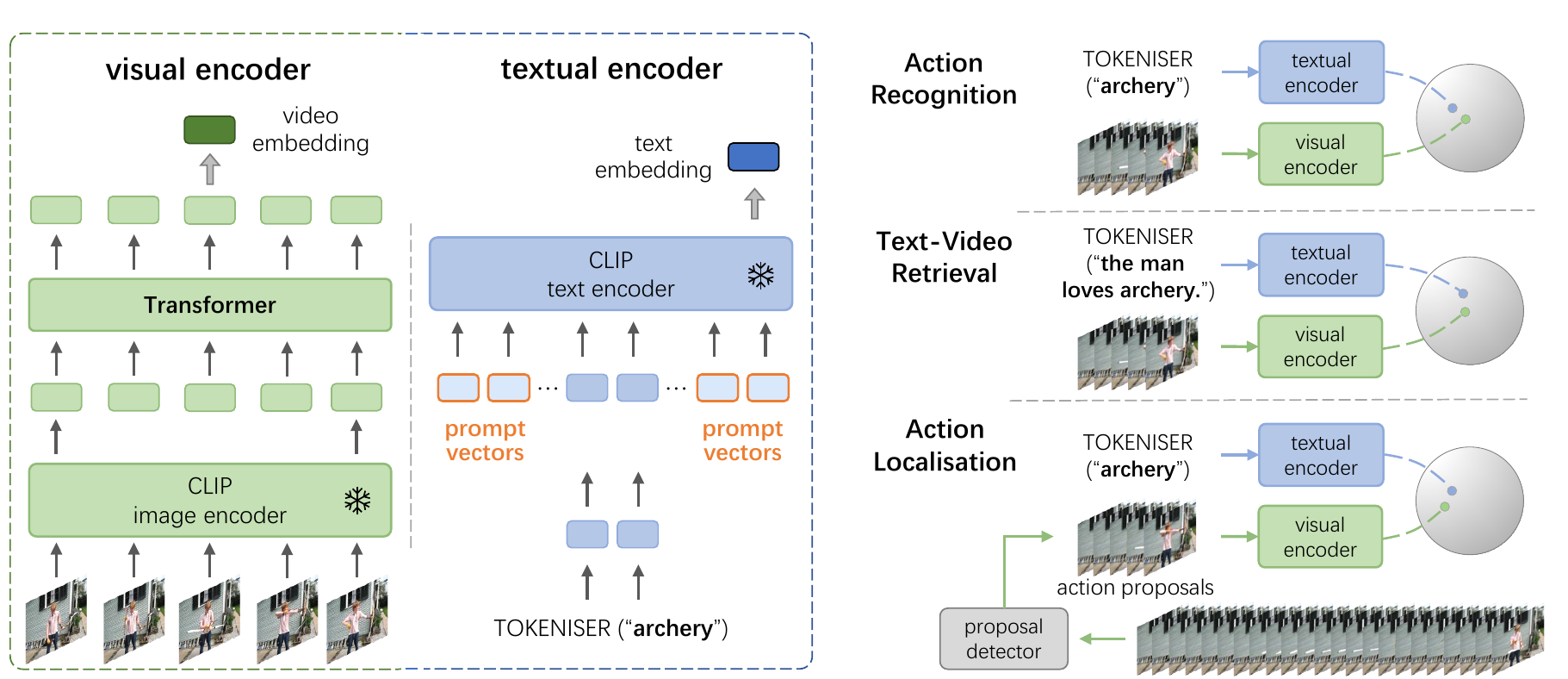} 
\end{center}   
\vspace{-0.3cm}
\caption{\small
{\bf Framework Overview}. We prepend\,/\,append several learnable prompt vectors to the CLIP text encoder to generate action classifiers or query embeddings; adopt a lightweight Transformer on the top of the CLIP image encoder for temporal modeling. During training, both the image and text encoders are kept {\em frozen}. By optimising the task-specific prompt vectors and temporal Transformer, we efficiently adapt CLIP to various video understanding tasks: action recognition, text-video retrieval, and action localisation, across closed-set, few-shot, and zero-shot scenarios.}
\label{fig:framework}
\end{figure*}
% ------------------------------------------------------
% ------------------------------------------------------

\section{Method} \label{sec:method}
Our goal is to efficiently steer a pre-trained \textbf{I}mage-based \textbf{V}isual-\textbf{L}anguage model (\textbf{I-VL}) to tackle novel downstream tasks, which we term as model adaptation. Here, we consider resource-hungry video understanding, {\em i.e.}~action recognition, action localisation, and text-video retrieval. To be self-contained, in Section~\ref{subsec:CLIP}, we briefly review the pre-training and inference of I-VL models; in Section~\ref{subsec:prompt}, we describe the proposed prompt learning and temporal modeling.

\subsection{Image-based Visual-Language Model} \label{subsec:CLIP}
\noindent {\bf {Pre-training}.}
\hspace{3.5pt} Given $N$ (image, text) pairs in one batch, the feature embeddings for image and text are computed with two individual encoders, and a dense cosine similarity matrix is calculated between all $N$ possible (image, text) pairs. The training objective is to jointly optimise the image and text encoders, by maximizing the similarity between $N$ correct pairs of (image, text) associations, while minimizing the similarity for $N \times (N-1)$ incorrect pairs by a symmetrical cross-entropy over the dense matrix, {\em i.e.}~noise contrastive learning.

Note that, both encoders contain a {\bf \textsc{tokeniser}} for converting image patches or language words to vectors. In particular, the input images are divided into patches and flattened into vectors, also called ``visual tokens''; while the input texts are converted into vectors~(``textual tokens'') by a trainable look-up table.

\vspace{0.15cm}
\par{\noindent \bf {Inference}.}
Once trained, the I-VL model can be deployed for image classification tasks on open vocabulary (zero-shot generalisation), with the corresponding visual classifiers being generated from the text encoder~$\mathrm{\Phi}_{\text{text}}$, which resembles the idea of hypernetwork~\cite{Ha16}. For example, to classify an image as cat or dog, the classifiers~($c_{\text{cat}}$ and $c_{\text{dog}}$) can be generated as:
\begin{align*}
c_{\text{cat}} &= \mathrm{\Phi}_{\text{text}}(\textsc{tokeniser}(\text{``this is a photo of [\underline{cat}]''})) \\
c_{\text{dog}} &= \mathrm{\Phi}_{\text{text}}(\textsc{tokeniser}(\text{``this is a photo of [\underline{dog}]''})) 
\end{align*}
and ``this is a photo of [$\cdot$]'' is a handcrafted prompt template, which has shown to be effective for image classification~\cite{Radford21}.

\vspace{0.15cm}
\par{\noindent \bf {Discussion}.}
Despite the tremendous success on zero-shot image classification, the I-VL model has also shown to be sensitive to the handcrafted prompt template, clearly posing limitations on its efficient adaptation for novel downstream tasks, where the expert knowledge might be difficult to condense or unavailable. Therefore, we consider to automate such prompt design procedures, exploring efficient approaches to adapt the pre-trained image-based visual-language model for novel downstream tasks, with minimal training.

% ------------------------------------------------------
% ------------------------------------------------------
% ---------------------------------------------------------------------
\subsection{Prompting CLIP for Video Understanding}  \label{subsec:prompt}
In general, we believe that prompt learning on I-VL models will shine in video domains for two main reasons: 1) Video tasks are resource-hungry. From the \textit{data} perspective, video-text pairs are harder to collect than image-text pairs. From the computation perspective, given the same budget, training on image-text pairs enables the model to learn more \textit{diversity}. Thus, it is more cost-effective to train large-scale I-VL models, and prompt them for efficient video understanding. 2) Videos are composed of frame sequences, establishing temporal dependencies on powerful image-based models is a natural and economical choice.

Next, we start by formulating the problem scenario and notations; then introduce the idea for efficient model adaptation through prompt learning; lastly, we augment the I-VL image encoder via temporal modeling, disambiguating the actions that require temporal reasoning. 
Among the I-VL models, CLIP~\cite{Radford21} is the publicly available milestone, we hence base this research on it, but the same technique should be applicable to other I-VL models as well.

\subsubsection{Problem Scenario.}  \label{subsec:problem_scenario}
\hspace{1pt} Given a video dataset that consists of training and validation sets, $\mathcal{D} = \{ \mathcal{D}_{\text{train}}, \mathcal{D}_{\text{val}} \}$, {\em e.g.}~$\mathcal{D}_{\text{train}} = \{(\mathcal{V}_1, y_1), \dots, (\mathcal{V}_n, y_n) \}$. The video $\mathcal{V}_i \in \mathbb{R}^{T \times H \times W \times 3}$ can range from seconds~(recognition and retrieval), to minutes long~(localisation). Respectively, $y_i$ either refers to {\em one} of the $\mathcal{C}_{\text{train}}$ action labels in the text format for recognition, {\em e.g.}~$y_i = \text{`archery'}$; or {\em dense} action category labels of $T$ timestamps for localisation, {\em e.g.}~$y_i \in \mathbb{R}^{T \times \mathcal{C}_{\text{train}}}$; or fine-grained text descriptions for retrieval, {\em e.g.}~$y_i = \text{`fry the onion in a pan'}$.

In the closed-set scenario, the action categories for training and evaluation are the same, {\em i.e.}~$\mathcal{C}_{\text{train}} = \mathcal{C}_{\text{val}}$; while in the zero-shot case, the action categories for training and evaluation are disjoint, {\em i.e.}~$\mathcal{C}_{\text{train}} \cap \mathcal{C}_{\text{val}} = \emptyset$.

% ---------------------------------------------------------------------
\subsubsection{Model Adaptation by Learning Prompts.} \label{subsec:textual}
The goal here is to steer pre-trained CLIP to perform various video tasks with minimal training. In specific, we strive for efficient model adaptation by prepending\,/\,appending a sequence of continuous random vectors (``prompt vectors'') with the textual tokens. While training, both the image and text encoders of CLIP are kept {\em frozen}, and the gradients will flow through the text encoder to only update the prompt vectors. Ultimately, these learnable vectors end up constructing ``virtual'' prompt templates that can be understood by the text encoder, and to generate desired classifiers or query embeddings, as detailed below.

\vspace{0.15cm}
\noindent {\bf (a) Action Recognition}
considers to classify the video clip or snippet into one of action categories. To generate the action classifier, we construct the ``virtual'' prompt template via feeding the tokenised category name into the pre-trained text encoder~$\mathrm{\Phi}_{\text{text}}$, for instance:
\begin{align*}
&c_{\text{archery}} = \mathrm{\Phi}_{\text{text}}(a_{1}, \dots, \textsc{tokeniser}(\text{``\underline{archery}''}), \dots, a_{k}) \\
&c_{\text{bowling}} = \mathrm{\Phi}_{\text{text}}(a_{1}, \dots,  \textsc{tokeniser}(\text{``\underline{bowling}''}), \dots, a_{k}) 
\end{align*}
where $a_i \in \mathbb{R} ^{D}$ denotes the $i$-th prompt vector, consisting of several learnable parameters, and $D$ is the vector dimension. $c_{\text{archery}}$ refers to the generated classifier for the action of ``archery''. Note that, the prompt vectors $\{a_i\}$ are shared for all action categories, thus they are only task-specific.

\vspace{0.15cm}
\noindent {\bf (b) Action Localisation} 
considers to localise and classify actions in untrimmed long videos. Here, we adopt the two-stage paradigm~\cite{Chao18,zhao2020bottom} to first detect potential class-agnostic action proposals~(detailed in Section~\ref{subsec:implementation details}), and followed by performing action classification on these detected proposals.

\vspace{0.15cm}
\noindent {\bf (c) Text-Video Retrieval}
considers to jointly learn visual and textual embeddings that pair the video and its corresponding textual description. In contrast to action recognition, where a video snippet is coarsely labeled by an action category, the text description in video retrieval contains more fine-grained details, usually a sentence. We here similarly \textsc{tokenise} the entire sentence, and feed the tokenised results to the text encoder with learnable prompt vectors, to generate the {\em query embedding} for each sentence.

\vspace{0.15cm}
\noindent {\bf (d) Summary.}
Generally speaking, learning prompts for model adaptation offers the following benefits: 
1) As both classification and retrieval can be tackled with one framework, with classifiers or query embeddings generated from text, either category names or free-form descriptions, all tasks can utilise {\em one} shared backbone, yet achieve competitive performance~(Section~\ref{sec:experiments}); 
2) Adapting to novel tasks only requires to optimise a few prompt vectors, facilitating the few-shot problem~(Table~\ref{tab:fewshot}); 
3) It enables to make better use of abundant training data, and further generalise beyond the closed-set categories~(Table~\ref{tab:vp} \&~\ref{tab:openset_localisation}).

% ------------------------------------------------------------------
\subsubsection{Temporal Modeling.} \label{subsec:visual}
As for pre-training, CLIP has thoroughly relied on the (image, text) pairs, posing clear pros on cons. On the one hand, the training data can be easily crawled from the web, which enables to learn much richer contents under a given compute constraint. However, on the other hand, it ignores the temporal component of the visual scene, and struggles to recognise the dynamic events, {\em e.g.}~push or pull, open or close. In this section, we bridge this image-to-video gap with a simple and lightweight temporal modeling module.

\vspace{0.05cm}
To be specific, we upgrade the CLIP image encoder $\mathrm{\Phi}_{\text{image}}$ into a video one $\mathrm{\Phi}_{\text{video}}$, by attaching a Transformer Encoder on top of frame-wise features from the frozen image encoder:
\begin{align*}
v_i = \mathrm{\Phi}_{\text{video}}(\mathcal{V}_i) = \mathrm{\Phi}_{\textsc{temp}}(\{ \mathrm{\Phi}_{\text{image}}(I_{i1}), \dots, \mathrm{\Phi}_{\text{image}}(I_{iT})\})
\end{align*}
where $\mathrm{\Phi}_{\textsc{temp}}$ refers to temporal modeling module, which is a multi-layer Transformer Encoder, consisting of Multi-head Self-attention, Layer Norm, and MLPs. To indicate the temporal order, we also add learnable temporal positional encoding onto image features. $v_i \in \mathbb{R} ^{T \times D}$ is dense feature embeddings of $T$ frames.

% -----------------------------------------
\subsubsection{Training Loss.}  \label{subsec:loss}
\hspace{3pt} Given a batch of (video, text) training pairs, the visual stream ends up with dense frame-wise feature embeddings~($v_i$); while for the textual stream, depending on the considered downstream tasks, it ends up with a set of action classifiers~($c_i \in \mathcal{C}_{\text{action}}$) or textual query embeddings~($c_i \in \mathcal{C}_{\text{query}}$).

For action recognition and text-video retrieval, we further compute the video-snippet-level feature by taking the mean pooling of the dense features:
\begin{align}
    \overline{v}_i = \mathrm{\Phi}_{\textsc{pool}}(v_i) \in \mathbb{R}^{1 \times D}
\end{align}

For action localisation, we take the mean pooling of the dense features within each detected action proposal, to obtain the proposal-level feature. And for simplicity, we also denote this proposal-level feature as $\overline{v}_i$.

During training, we jointly optimise the textual prompt vectors and temporal Transformer, such that the video snippet (proposal) features and its paired classifier or textual query embedding emit the highest similarity score among others. This is achieved with a simple NCE loss:
\begin{align}
\mathcal{L} = &- \sum_i \big( \log \frac{\exp(\overline{v}_i \cdot c_{i} / \tau)}{\sum\limits_{j} \exp(\overline{v}_i \cdot c_j / \tau)} \big)
\end{align}

Note that, both $\overline{v}_i$ and $c_j$ have been L2-normalised here, and $\tau$ refers to the temperature hyper-parameter for scaling. In this way, various video tasks are formulated under the same umbrella, we therefore effectively close the optimisation objective gap between CLIP pre-training and video understanding.

% ------------------------------------------------------
% ------------------------------------------------------
% ------------------------------------------------------
\section{Experiments}  \label{sec:experiments}
We experiment $3$ fundamental video tasks, across $10$ standard datasets. In Section~\ref{sec:recognition experiments}, we conduct ablation studies on action recognition, to validate the effectiveness of prompt learning and temporal modeling. In Section~\ref{sec:localisation experiments} and~\ref{sec:retrieval experiments}, we further benchmark two other popular tasks: action localisation and text retrieval.

\subsection{Implementation Details} \label{subsec:implementation details}
The image and text encoders are adopted from pre-trained CLIP (ViT-B/16), and are both kept frozen. The only trainable parameters are prompt vectors and temporal Transformer, which have the same dimension ($D = 512$), and are both initialised with $\mathcal{N}(0,0.01)$. All video frames are pre-processed to $224 \times 224$ spatial resolution, the maximum number of textual tokens is 77~(follow the official CLIP design), and the temperature hyper-parameter $\tau$ is set to $0.07$.

For action recognition, all the videos are decoded to $30$ fps, and each video is sampled $16$ frames with a random frame gap~($\text{gap} \in \{1,2,3,4,5,6,10,15\}$) for training~\cite{Wang16-TSN}. The temporal positional encodings consist of each frame's index and the frame sampling gap (video playing speed). The model is optimised using AdamW~\cite{loshchilov2017decoupled} with a learning rate of $10^{-4}$, and a batch size of $64$ videos. During inference, we perform $5$-crop evaluation, {\em i.e.}~random sample $16$ frames from each video for $5$ times, and take the average results as the final predictions.

For action localisation, we follow the two-stage paradigm: class-agnostic proposal detection and proposal classification. To generate high-quality action proposals, we first divide the entire video into several equal-frame segments; then use the CLIP image encoder with one Transformer layer to extract frame-wise embeddings; and finally feed these embeddings to the off-the-shelf proposal detectors~\cite{lin2021learning,yang2020revisiting}. These detectors build feature pyramids, and make predictions in parallel to determine the actionness, centerness, and boundaries. Note that, our method is flexible to the choice of proposal detectors, and we do not innovate on such candidate proposal procedures. To generate proposal classifiers, we adopt the same implementation details as for action recognition.

For video retrieval, we take the $16$-frame input with a random frame gap~($\text{gap} \in \{10,15,30\}$). Note that, here we use significantly larger frame gaps than action recognition, as retrieval tends to require long-term visual dependencies.

\begin{table*}[t]
\scriptsize
\centering
\caption{\footnotesize \textbf{Ablation study for closed-set action recognition}. TFM is the number of Transformer layers for temporal modeling. Baseline-I denotes the ``zero-shot'' CLIP inference with the handcrafted prompt template (``a photo of [$\cdot$].''). Baseline-II is the standard practice for training linear probe on the CLIP image encoder.}
\begin{tabular}[t]{C{2.15cm}C{1.6cm}C{1.6cm}|C{0.9cm}C{0.9cm}C{0.9cm}|C{0.9cm}C{0.9cm}C{0.9cm}}
\toprule
& & & \multicolumn{3}{c}{K-400} & \multicolumn{3}{c}{K-700} \\ \toprule
Model & Prompt & Temporal & TOP1 & TOP5 & AVG & TOP1 & TOP5 & AVG   \\ \midrule
Baseline-I~\cite{Radford21} & \text{hand-craft}  & \xmark & -- & -- & -- & -- & -- & 52.4 \\ 
Baseline-II~\cite{Radford21} & \xmark  & \xmark & -- & -- & -- & -- & -- & 66.1 \\ \midrule
A0 & 2+X+2   & \xmark & 65.4 & 88.7 & 77.1 & 56.3 & 81.9 & 69.1 \\ 
A1 & 4+X+4   & \xmark & 66.1 & 89.0 & 77.6 & 56.6 & 82.4 & 69.5 \\
A2 & 8+X+8   & \xmark & 67.9 & 90.0 & 79.0 & 57.4 & 83.0 & 70.2 \\
A3 & 16+X+16 & \xmark & 68.8 & 90.1 & 79.5 & 57.8 & 83.1 & 70.5\\ \midrule
A4 & 16+X+16 & 1-TFM  &  75.8 & 92.9 & 84.4 & 64.2 & 87.3 & 75.8 \\ 
A5 & 16+X+16 & 2-TFM  & 76.6 & 93.3 & 85.0 & 64.7 & 88.5 & 76.6 \\ 
A6 & 16+X+16 & 3-TFM  & 76.9 & 93.5 & 85.2 & 64.8 & 88.4 & 76.6 \\ 
A7 & 16+X+16 & 4-TFM  & 76.8 & 93.5 & 85.2 & 64.9 & 87.9 & 76.4 \\ \bottomrule
\end{tabular}
\label{tab:closet_ablation}
\end{table*}

\subsection{Action Recognition} \label{sec:recognition experiments}
\par{\noindent \bf Datasets \& Metrics.}
{\bf HMDB-51}~\cite{Kuehne11} contains 7k videos of $51$ categories. Its standard split is to train on $3570$ videos, and evaluate on another $1530$ videos. {\bf UCF-101}~\cite{Soomro12} covers 13k videos spanning $101$ categories. The standard split is to train on $9537$ videos and evaluate on the left $3783$ videos. {\bf Kinetics-400}~\cite{kay2017} (K-400) covers around 230k $10$-second video clips sourced from YouTube. {\bf Kinetics-700}~\cite{Carreira19} (K-700) is simply an extension of K-400, with around 650k video clips. {\bf Something-Something} V2~\cite{goyal2017something} covers $174$ action categories. Its standard split is $168,913$ training videos, $24,777$ validation videos, and $27,157$ testing videos. Some categories are fine-grained, {\em e.g.} bending something so that it deforms {\em v.s} bending something until it breaks. For evaluation metrics, we report the standard TOP1 and TOP5 accuracy, and the average of these two metrics.

\subsubsection{Closed-set Action Recognition} is the common scenario, where the model is trained and evaluated on videos from the same categories, {\em i.e.} $\mathcal{C}_{\text{train}} = \mathcal{C}_{\text{val}}$. For comprehensive comparisons, we here adopt the standard splits of four popular benchmarks, namely, HMDB-51, UCF-101, K-400, and K-700.

\vspace{0.15cm}
\noindent $\bullet$ \textit{Ablation Studies} are conducted on two largest benchmarks. Table~\ref{tab:closet_ablation} presents the results for prompt learning and temporal modeling. The prompt here follows the format of $[a_1,..\,, a_k, \mathrm{X}, a_{k+1}, ..\,, a_{2k}]$. Note that, although we prepend and append the equal number of prompt vectors, the optimisation can perfectly learn to ignore any of these vectors, thus, we do not ablate other prompt formats.

As the baselines, we compare with the official results reported in the original CLIP~\cite{Radford21}. Specifically, Baseline-I refers to the ``zero-shot'' inference with handcrafted prompt templates (``a photo of [$\cdot$].''), and Baseline-II denotes the standard practice for training linear classifiers on top of the pre-trained CLIP image encoder with the considered downstream datasets.

Generally speaking, training more text prompt vectors brings consistent improvements on both TOP1 and TOP5 accuracy; In addition, adding temporal modeling also brings immediate benefits, with average gains of 4.9\% and 5.3\% on K-400 and K-700. However, it gives diminishing returns as more Transformer layers are added. Overall, all the results suggest that, both the prompt learning and temporal modeling are essential. While comparing with Baseline-I, the A3 model demonstrates a performance boost of 18.1\%, clearly showing the benefits of learning prompt vectors over the handcrafted ones. Moreover, even with fewer trainable parameters (only 16K), the A3 model also surpasses Baseline-II, with 4.4\% gains, showing the superiority of prompting adaptation.

\vspace{0.03cm}
For all the following action recognition experiments, we inherit the best practice from the ablation studies, {\em i.e.} prepend\,/\,append $16$ prompt vectors to category names, and only use two Transformer layers (5M parameters) for temporal modeling, for its best trade-off on performance and computational cost.

% ==================================================
\begin{table*}[t]
\centering
\scriptsize
\caption{\footnotesize \textbf{Comparison on closed-set action recognition}. On all datasets, our model performs comparably to existing methods, by training far fewer parameters.}
\begin{tabular}[t]{C{2.6cm}|C{1.0cm}C{1.05cm}|C{1.0cm}C{1.0cm}|C{1.0cm}C{1.05cm}|C{1.0cm}C{1.0cm}}
\toprule
& \multicolumn{2}{c}{HMDB-51} & \multicolumn{2}{c}{UCF-101} & \multicolumn{2}{c}{K-400} & \multicolumn{2}{c}{K-700}  \\ \midrule
Method & TOP1 & TOP5 & TOP1 & TOP5 & TOP1 & TOP5 & TOP1 & TOP5\\ \midrule
I3D~\cite{carreira2017quo} & 74.3 & -- & 95.1 & -- & 71.6 & 90.0 & 58.7 &  81.7 \\
S3D-G~\cite{Xie18-S3D} & 75.9 & -- & 96.8 & -- & 74.7 & 93.4 & -- &  -- \\
R(2+1)D~\cite{Tran18} & 74.5 & -- & 96.8 & -- & 72.0 & 90.0 & -- &  -- \\
TSM~\cite{lin2019tsm} & -- & -- & -- & -- & 74.7 & -- & -- &  -- \\
R3D-50~\cite{hara2018can} & 66.0 & -- & 92.0 & -- & -- & -- & 54.7 & -- \\
NL-I3D~\cite{wang2018non} & 66.0 & -- & -- & -- & 76.5 & 92.6 & -- & -- \\
SlowFast~\cite{Feichtenhofer19} & -- & -- & -- & -- & 77.0 & 92.6 & -- &  -- \\
X3D-XXL~\cite{Feichtenhofer20} & -- & -- & -- & -- & 80.4 & 94.6 & -- &  -- \\
TimeSformer-L~\cite{Bertasius21} & -- & -- & -- & -- & 80.7 & 94.7 & -- &  -- \\ \midrule
{Ours~(A5)} & 66.4 & 92.1 & 93.6 & 99.0 & 76.6 & 93.3 & 64.7 & 88.5 \\ \bottomrule
\end{tabular}
\vspace{0.2cm}
\label{tab:closet_SOTA}
\end{table*}
% ==================================================

\begin{table*}[t]
\scriptsize
\centering
\parbox{.428\textwidth}{\caption{\footnotesize \textbf{Something-Something V2 closed-set action recognition}. Baseline-I refers to CLIP ``zero-shot'' inference, without using any prompt template. TFM refers to the number of temporal Transformer layers.}}
\hspace{\fill}
\parbox{.55\textwidth}{\begin{tabular}[t]{C{2.1cm}C{1.4cm}C{1.4cm}|C{1.1cm}}
\toprule
Method & Prompt & Temporal & TOP1   \\ \midrule
TRN~\cite{zhou2018temporal} & \xmark & \xmark & 48.8 \\ 
SlowFast~\cite{feichtenhofer2020x3d} & \xmark & \xmark & 61.7  \\ 
TSM~\cite{lin2019tsm} & \xmark & \xmark & 63.4 \\ 
ViVIT~\cite{arnab2021vivit} & \xmark & \xmark & 65.9 \\ 
Swin-B~\cite{liu2022video} & \xmark & \xmark & 69.6 \\ \midrule
Baseline-I~\cite{Radford21} & \xmark & \xmark & 1.4 \\ 
B1 & 16+X+16 & \xmark & 18.4 \\ 
B2 & 16+X+16 & 4-TFM  & 38.1 \\ 
\bottomrule
\end{tabular}}
\label{tab:sth-sth}
\end{table*}

% ==================================================
\begin{table}[t]
\scriptsize
\centering
\caption{\footnotesize \textbf{Comparison on few-shot action recognition}. Here, $\mathcal{C}_{\text{ALL}}$ refers to the case where the model is evaluated on all action categories of the corresponding dataset, rather than only 5-way classification, {\em e.g.}~101 categories for UCF, 400 categories for K-400. Baseline-I denotes the ``zero-shot'' CLIP inference with handcrafted prompts.}
\begin{tabular}[t]{C{2cm}|C{2.1cm}|C{1.45cm}C{1.45cm}|C{1.35cm}C{1.35cm}C{1.2cm}}
\toprule
Method & K-shot N-way & Prompt & Temporal & UCF-101 & HMDB-51 & K-400 \\ \midrule
CMN~\cite{Zhu18}  & 5 \qquad 5 & -- & -- & -- & -- & 78.9 \\
TARN~\cite{bishay2019tarn} & 5 \qquad 5 & -- & -- & -- & -- & 78.5 \\ 
ARN~\cite{Zhang20a} & 5 \qquad 5 & -- & -- & 83.1 & 60.6 & 82.4  \\ 
TRX~\cite{Perrett21} & 5 \qquad 5 & -- & -- & 96.1 & 75.6 & 85.9 \\ \midrule
Baseline-I~\cite{Radford21} & -- \qquad 5 & hand-craft & \xmark & 91.9 & 68.9 & 95.1 \\
\multirow{2}{*}{Ours} & 5 \qquad 5 & \checkmark & \xmark & {\bf 98.3} & {\bf 85.3} & {\bf 96.4} \\
& 5 \qquad 5 & \checkmark & \checkmark & 97.8 & 84.9 & 96.0 \\ \midrule
Baseline-I~\cite{Radford21} & \ -- \quad \; $\mathcal{C}_{\text{ALL}}$ & hand-craft & \xmark & 64.7 & 40.1 & 54.2 \\
\multirow{2}{*}{Ours} & \ 5 \quad \; $\mathcal{C}_{\text{ALL}}$ & \checkmark & \xmark & 77.6 & 56.0 & 57.1 \\
 & \ 5 \quad \;  $\mathcal{C}_{\text{ALL}}$ & \checkmark & \checkmark & {\bf 79.5} & {\bf 56.6} & {\bf 58.5} \\ \bottomrule
\end{tabular}
\label{tab:fewshot}
\end{table}
% ==================================================

\vspace{0.15cm}
\noindent $\bullet$ \textit{Comparison to SOTA}. 
\hspace{1pt} Table~\ref{tab:closet_SOTA} compares our method with existing state-of-the-art approaches on four popular action recognition benchmarks. Overall, on all datasets, our model performs comparably with the competitors, although we only need to train {\em far fewer} parameters (around 5M), {\em i.e.}~two Transformer layers and several prompt vectors, advocating efficient model adaptation.

Table 3 further explores our method on the fine-grained motion benchmark: Something-Something. Comparing to the simple Baseline-I, {\em i.e.} the ``zero-shot'' CLIP inference without any prompt, both prompt learning and temporal Transformer bring considerable performance gains. However, there is still a certain gap between our results and existing state-of-the-art methods, we conjecture that this may be due to that, the CLIP pre-training relies more on object information for action recognition, lacking prior knowledge of fine-grained motions.

\subsubsection{Few-shot Action Recognition} \label{subsubsec:few-shot action recognition}
aims to classify videos with only a few training samples, in this section, we benchmark on two different settings. The first one follows the previous literature~\cite{Cao20,Perrett21,Zhang20a}, and evaluates on the standard $K$-shot, $N$-way classification; while the second part considers a more challenging setting that classifies all categories with $K$-shot support samples. For more details on dataset splits, please refer to the Appendix. In both settings, we use the ``zero-shot'' CLIP inference with handcrafted templates as the baseline.

\vspace{0.2cm}
\noindent $\bullet$ \textit{$5$-Shot-$5$-Way Setting}.
\hspace{1pt} For fair comparisons, this setting adopts the publicly accessible few-shot splits. Specifically, for HMDB-51 and UCF-101, we follow~\cite{Zhang20a} to collect $10$ and $21$ testing action categories respectively; while for K-400, we follow~\cite{Perrett21,Zhu18} to collect $24$ testing categories, each containing $100$ videos. During training, we sample $5$ categories (ways) from the above data, with $5$ videos (shots) from each category, and utilise the remaining data for evaluation. To ensure the statistical significance, we conduct $200$ trials with random samplings.

\vspace{0.03cm}
Table~\ref{tab:fewshot} presents the average TOP1 accuracy for three datasets. Our method (with/without temporal modeling) clearly outperforms all previous methods by a significant margin, around 10\% on HMDB-51 and K-400, demonstrating the superiority of our proposed idea for model adaptation.

\vspace{0.2cm}
\noindent $\bullet$ \textit{$5$-Shot-$\mathcal{C}$-Way Setting}.
\hspace{1pt} Here, we further consider a more challenging scenario: scaling the problem up to classifying all action categories in the dataset with only 5 training samples per category, for example, $\mathcal{C}_{\text{ALL}}=400$ for K-400, $\mathcal{C}_{\text{ALL}}=101$ for UCF-101. Specifically, on each dataset, we sample $5$ videos (shots) from the training set for each category, to form the few-shot support set, and then measure performance on the corresponding standard testing set.

For this experiment setting, we conduct $10$ random sampling rounds, and also record the average TOP1 accuracy in Table~\ref{tab:fewshot}. Comparing to the $5$-way classification, the $\mathcal{C}$-way setting is clearly more challenging, our model (with/without temporal modeling) still shows promising results. While comparing to the Baseline-I, our performance gains on both UCF-101 and HMDB-51 are around 15\%.

% ==================================================
\begin{table}[t]
\setlength\tabcolsep{5pt}
\scriptsize
\caption{\footnotesize \textbf{Ablation study for zero-shot action recognition on K-700}. Baseline-I refers to the results from the CLIP zero-shot evaluation. The model is trained on 400 action categories and evaluated on the other 300 disjoint categories.}
\centering
\begin{tabular}[t]{C{2cm}C{2cm}C{1.8cm}|C{1.2cm}C{1cm}C{1cm}}
\toprule
Model & Prompt & Temporal & TOP1 & TOP5  & AVG \\ \midrule
Baseline-I~\cite{Radford21} & hand-craft & \xmark & 52.4 & 77.3 & 64.9\\ \midrule
C0 & 4+X+4     & \xmark & 57.4 & 83.3 & 70.4\\ 
C1 & 8+X+8     & \xmark & 57.7 & 82.6 & 70.2\\ 
C2 & 16+X+16   & \xmark & {\bf 58.4} & 82.6  & 70.5\\ 
C3 & 32+X+32   & \xmark & 57.5 & {\bf 84.6} & {\bf 71.1}\\  \midrule
C4 & 16+X+16   & 1-TFM  & 47.9 & 76.8 & 62.4\\ 
C5 & 16+X+16   & 2-TFM  & 45.5 & 75.4 & 60.5\\ 
C6 & 16+X+16   & 3-TFM  & 45.6 & 75.2 & 60.4\\  \bottomrule
\end{tabular}
\label{tab:vp}
\end{table}
% ==================================================

% ----------------------------------------------------------------------------
\subsubsection{Zero-shot Action Recognition} refers to the novel scenario, where videos for training and validation are from different action categories, {\em i.e.}~$\mathcal{C}_{\text{train}} \cap \mathcal{C}_{\text{val}} = \emptyset$. Specifically, we split K-700 into two parts, with $\mathcal{C}_{\text{train}} = 400$ categories for training, and the remaining $\mathcal{C}_{\text{val}} = 300$ categories for evaluation. For more details on dataset splits, please refer to the Appendix.

\vspace{0.03cm}
As a baseline, we evaluate the CLIP with handcrafted prompt templates. As reported in Table~\ref{tab:vp}, our model achieves 6.0\% gains on TOP1 accuracy over the Baseline-I, showing the effectiveness of prompt learning for zero-shot recognition. Interestingly, the number of learnable prompt vectors does not make a difference, and adding temporal modeling diminishes the performance gain. We conjecture this is because the additional Transformer layer could specialise on the training set, thus harming the generalisation towards unseen action categories.

\subsubsection{Conclusion \& Discussion.}
Among all benchmarks on action recognition, we have demonstrated the effectiveness of prompt learning and temporal modeling. For closed-set recognition, even without temporal modeling, learnable prompts clearly surpass the handcrafted ones, and linear probe settings. While comparing to state-of-the-art approaches, despite training {\em far fewer} parameters, our model still shows competitive performance on all benchmarks. For few-shot recognition with limited training samples, model adaptation through prompt learning really shines, outperforming all previous methods significantly. Lastly, for zero-shot scenarios, textual prompts enable to make better use of abundant training data, and further improve the generalisation beyond the seen categories.

% ==================================================
\begin{table*}[t]
\scriptsize
\centering
\caption{\footnotesize \textbf{Comparison on closed-set action localisation}. AVG denotes the average mAP in [0.3:0.1:0.7] on THUMOS14, and [0.5:0.05:0.95] on ActivityNet1.3. Baseline-III adopts the same first-stage proposal detector as our method, but uses the original CLIP with handcrafted prompts as the second-stage proposal classifier.}
\begin{tabular}{C{2.0cm}C{0.85cm}C{1.65cm}|C{0.65cm}C{0.6cm}C{0.6cm}C{0.6cm}C{0.6cm}C{0.65cm}|C{0.65cm}C{0.6cm}C{0.6cm}C{0.65cm}} 
\toprule 
\multirow{2}{*}{} & \multirow{2}{*}{} & \multirow{2}{*}{} & \multicolumn{6}{c}{THUMOS14} & \multicolumn{4}{c}{ActivityNet1.3} \\  \midrule
Method & Date & Modality & 0.3 & 0.4 & 0.5 & 0.6 & 0.7 & AVG & 0.5 & 0.75 & 0.95 & AVG \\ \midrule
\multicolumn{1}{c|}{CDC~\cite{shou2017cdc}} & \multicolumn{1}{c|}{2017} & \multicolumn{1}{c|}{RGB+Flow} & 40.1 & 29.4 & 23.3 & 13.1 & 7.9 & 22.8 & 45.3 & 26.0 & 0.2 & 23.8 \\
\multicolumn{1}{c|}{TALNET~\cite{Chao18}} & \multicolumn{1}{c|}{2018} & \multicolumn{1}{c|}{RGB+Flow} & 53.2 & 48.5 & 42.8 & 33.8 & 20.8 & 39.8 & 38.2 & 18.3 & 1.3 & 20.2 \\
\multicolumn{1}{c|}{BSN~\cite{Lin18}} & \multicolumn{1}{c|}{2018} & \multicolumn{1}{c|}{RGB+Flow} & 53.5 & 45.0 & 36.9 & 28.4 & 20.0 & 36.8 & 46.5 & 30.0 & 8.0 & 30.0 \\
\multicolumn{1}{c|}{DBS~\cite{gao2019video}} & \multicolumn{1}{c|}{2019} & \multicolumn{1}{c|}{RGB+Flow} & 50.6 & 43.1 & 34.3 & 24.4 & 14.7 & 33.4 & -- & -- & -- & --\\
\multicolumn{1}{c|}{BUTAL~\cite{zhao2020bottom}} & \multicolumn{1}{c|}{2020} & \multicolumn{1}{c|}{RGB+Flow} & 53.9 & 50.7 & 45.4 & 38.0 & 28.5 & 43.3 & 43.5 & 33.9 & {\bf 9.2} & 30.1 \\
\multicolumn{1}{c|}{A2NET~\cite{yang2020revisiting}} & \multicolumn{1}{c|}{2020} & \multicolumn{1}{c|}{RGB+Flow} & 58.6 & 54.1 & 45.5 & 32.5 & 17.2 & 41.6 & 43.6 & 28.7 & 3.7 & 27.8 \\
\multicolumn{1}{c|}{GTAD~\cite{xu2020g}} & \multicolumn{1}{c|}{2020} & \multicolumn{1}{c|}{RGB+Flow} & 66.4 & 60.4 & 51.6 & 37.6 & 22.9 & 47.8 & 50.4 & 34.6 & 9.0 & 34.1 \\
\multicolumn{1}{c|}{BSN++~\cite{su2020bsn++}} & \multicolumn{1}{c|}{2021} & \multicolumn{1}{c|}{RGB+Flow} & 59.9 & 49.5 & 41.3 & 31.9 & 22.8 & 41.1 & 51.3 & {\bf 35.7} & 8.3 & {\bf 34.9}\\
\multicolumn{1}{c|}{AFSD~\cite{lin2021learning}} & \multicolumn{1}{c|}{2021} & \multicolumn{1}{c|}{RGB+Flow} & {\bf 67.3} & {\bf 62.4} & {\bf 55.5} & {\bf 43.7} & {\bf 31.1} & {\bf 52.0} & {\bf 52.4} & 35.3 & 6.5 & 34.4\\ \midrule
\multicolumn{1}{c|}{TALNET~\cite{Chao18}} & \multicolumn{1}{c|}{2018} & \multicolumn{1}{c|}{RGB} & 42.6 & -- & 31.9 & -- & 14.2 & --  & -- & -- & -- & -- \\
\multicolumn{1}{c|}{A2NET~\cite{yang2020revisiting}} & \multicolumn{1}{c|}{2020} & \multicolumn{1}{c|}{RGB} & 45.0 & 40.5 & 31.3 & 19.9 & 10.0 & 29.3 & 39.6 & 25.7 & 2.8 & 24.8 \\
\multicolumn{1}{c|}{Baseline-III} & \multicolumn{1}{c|}{2022} & \multicolumn{1}{c|}{RGB} & 36.3 & 31.9 & 25.4 & 17.8 & 10.4 & 24.3 & 28.2 & 18.3 & 3.7 & 18.2 \\
\multicolumn{1}{c|}{Ours} & \multicolumn{1}{c|}{2022} & \multicolumn{1}{c|}{RGB} & {\bf 50.8} & {\bf 44.1} & {\bf 35.8} & {\bf 25.7} & {\bf 15.7} & {\bf 34.5} & {\bf 44.0} & {\bf 27.0} & {\bf 5.1} & {\bf 27.3} \\
\bottomrule
\end{tabular}
\label{tab:closet_localisation}
\end{table*}

\subsection{Action Localisation} \label{sec:localisation experiments}
\noindent {\bf Datasets \& Metrics.}
{\bf THUMOS14}~\cite{THUMOS14} covers $413$ untrimmed sports videos from $20$ action categories, with an average of $15$ instances per video. The standard split is $200$ training videos and $213$ validation videos. {\bf ActivityNet1.3}~\cite{Caba15} has around 20k untrimmed videos of $200$ action categories. The standard split is $10,024$ training videos and $4,926$ validation videos. We evaluate with the mean Average Precision (mAP) at various IoU thresholds. On THUMOS14, the IoU set is $[0.3:0.1:0.7]$; as for ActivityNet1.3, the IoU set is $[0.5:0.05:0.95]$.

\subsubsection{Closed-set Action Localisation} is the commonly adopted setting, where the model is trained and tested on videos of the same categories, {\em i.e.}~$\mathcal{C}_{\text{train}} = \mathcal{C}_{\text{val}}$. For fair comparisons, we use the standard dataset splits as in the literature.

\vspace{0.03cm}
Table~\ref{tab:closet_localisation} reports the results. As a baseline, we adopt the same first-stage proposal detector, but utilise the original CLIP with handcrafted prompts (``this is an action of [$\cdot$]'') for the second-stage proposal classifier. On both datasets, our model significantly outperforms the Baseline-III, again confirming the effectiveness of prompt learning and temporal modeling. While comparing to other existing methods that use pre-trained RGB stream, our method also demonstrates superior performance, with around 5.2\% and 2.5\% gains on average mAP.

\subsubsection{Zero-shot Action Localisation} refers to the novel scenario, where the action categories for training and testing are disjoint. As we are not aware of any existing benchmarks on this challenging scenario, we initiate two evaluation settings on THUMOS14 and ActivityNet1.3: one is to train with 75\% categories and test on the left 25\% categories; the other is to train with 50\% categories and test on the remaining 50\% categories. To ensure the statistical significance, we conduct $10$ random samplings to split action categories for each setting.

% ==================================================
\begin{table}[t]
\scriptsize
\caption{\footnotesize \textbf{Results of zero-shot action localisation}. Baseline-III uses the same proposal detector as our method, but adopts the original CLIP with handcrafted prompts as the proposal classifier. Our model is trained on 75\% (or 50\%) action categories and tested on the remaining 25\% (or 50\%) action categories.}
\setlength\tabcolsep{3pt}
\centering
\begin{tabular}[t]{C{1.7cm}C{2cm}|C{0.6cm}C{0.5cm}C{0.5cm}C{0.5cm}C{0.5cm}C{0.55cm}|C{0.6cm}ccc}
\toprule
& \multirow{2}{*}{} & \multicolumn{6}{c}{THUMOS14} & \multicolumn{4}{c}{ActivityNet1.3} \\ \midrule
Method & Train {\em v.s}\, Test & 0.3 & 0.4 & 0.5 & 0.6 & 0.7 & AVG & 0.5 & 0.75 & 0.95 & AVG \\ \midrule
\multicolumn{1}{c|}{Baseline-III} & 75\% {\em v.s}\, 25\% & 33.0 & 25.5 & 18.3 & 11.6 & 5.7 & 18.8 & 35.6 & 20.4 & 2.1 & 20.2 \\
\multicolumn{1}{c|}{Ours} & 75\% {\em v.s}\, 25\% & {\bf 39.7} & {\bf 31.6} & {\bf 23.0} & {\bf 14.9} & {\bf 7.5} & {\bf 23.3} & {\bf 37.6} & {\bf 22.9} & {\bf 3.8} & {\bf 23.1} \\ \midrule
\multicolumn{1}{c|}{Baseline-III} &  50\% {\em v.s}\, 50\%  & 27.2 & 21.3 & 15.3 & 9.7 & 4.8 & 15.7 & 28.0 & 16.4 & 1.2 & 16.0 \\
\multicolumn{1}{c|}{Ours} & 50\% {\em v.s}\, 50\% & {\bf 37.2} & {\bf 29.6} & {\bf 21.6} & {\bf 14.0} & {\bf 7.2} & {\bf 21.9} & {\bf 32.0} & {\bf 19.3} & {\bf 2.9} & {\bf 19.6} \\ \bottomrule
\end{tabular}
\label{tab:openset_localisation}
\end{table}
% ==================================================

\vspace{0.03cm}
Table~\ref{tab:openset_localisation} shows the average performance. As proposals are class-agnostic, the key of two-stage localisation is the proposal classifier. For comparisons, we also implement the baseline, which uses the same proposal detector as our model, but classifies action proposals using original CLIP with handcrafted prompts. In both settings, our model shows superior performance than the Baseline-III. However, when comparing with closed-set, the zero-shot performance drops dramatically. Note that, such drop now comes from two sources: one is the recall drop from the first-stage class-agnostic proposals, because for zero-shot, there exist partial differences in action distributions of training and testing, causing proposals still be biased towards seen distributions; and the other comes from the second-stage classification errors. The complete ablations are available in the Appendix.

% ==================================================
\begin{table}[t]
\scriptsize
\centering
\caption{\footnotesize \textbf{Results of text-video retrieval}. Baseline-IV refers to the original CLIP model with text query na\"ively encoded, {\em i.e.}~without using any prompt. \textsc{E2E} denotes if the model has been trained end-to-end. As these methods are pre-trained on different datasets with variable sizes, it is unlikely to make fair comparisons.}
\setlength\tabcolsep{3pt}
\begin{tabular}[t]{C{2.0cm}C{0.7cm}|C{0.95cm}C{0.85cm}|C{0.85cm}C{0.75cm}|C{0.85cm}C{0.75cm}|C{0.85cm}C{0.75cm}}
\toprule
\multirow{2}{*}{} & & \multicolumn{2}{c}{MSRVTT~(9K)} & \multicolumn{2}{c}{LSMDC}
& \multicolumn{2}{c}{DiDeMo} & \multicolumn{2}{c}{SMIT} \\ \midrule
Method & \textsc{E2E} & R@1 & R@5 & R@1 & R@5 & R@1 & R@5 & R@1 & R@5 \\ \midrule
CE~\cite{liu2019use} & \xmark & 21.7 & 51.8 & 12.4 & 28.5  & 16.1 & 41.1 & -- & -- \\ 
MMT~\cite{gabeur2020multi} & \xmark & 24.6 & 54.0 & 13.2 & 29.2 & -- & -- & -- & -- \\ 
TT-CE+~\cite{croitoru2021teachtext} & \xmark & 29.6 & 61.6  & {\bf 17.2} & {\bf 36.5}  & 21.6 & 48.6 & -- & -- \\ \midrule
Baseline-IV & \xmark & 31.2 & 53.7  & 11.3 & 22.7  & 28.8 & 54.6 & 39.3 & 62.8 \\ 
Ours & \xmark & {\bf 36.7} & {\bf 64.6} & 13.4 & 29.5 &  {\bf 36.1} & {\bf 64.8}  &  {\bf 66.6} & {\bf 87.8} \\ \midrule \midrule
Frozen~\cite{bain2021frozen} & \checkmark & 31.0 & 59.5  & 15.0 & 30.8  & 34.6 & 65.0 & -- & -- \\
CLIP4Clip~\cite{Luo_CLIP4Clip21}  & \checkmark & 44.5 & 71.4 & 22.6 & 41.0  & 43.4 & 70.2 & -- & -- \\ \bottomrule
\end{tabular}
\label{tab:retrieval}
\end{table}

\subsection{Text-Video Retrieval}  \label{sec:retrieval experiments}
\noindent {\bf Datasets \& Metrics.}
{\bf MSRVTT}~\cite{xu2016msr} covers $10,000$ videos and $200,000$ captions. We train on ``Training-9K'' split~\cite{Gabeur20}, and test on ``test 1k-A''~\cite{yu2018joint} of $1,000$ clip-text pairs. {\bf LSMDC}~\cite{rohrbach2017movie} contains $118,081$ videos of $2$ to $30$ seconds. We train on $7,408$ validation videos, and evaluate on another $1,000$ videos. {\bf DiDeMo}~\cite{anne2017localizing} contains $10,464$ videos annotated with $40,543$ sentences. {\bf SMIT}~\cite{monfort2021spoken} contains more than 500k videos randomly chosen from M-MiT training set~\cite{monfort2021multi}, and 10k validation videos. Each video contains at least one text description. For evaluation metrics, we report the average recall at K (R@K). And the comprehensive results with median rank~(MdR) can be found in the Appendix.

\vspace{0.15cm}
\par{\noindent \bf Results.}
Table~\ref{tab:retrieval} presents text-retrieval results on four benchmarks. Note that, we here only employ $8$ learnable prompt vectors, {\em i.e.}~[4+X+4]. This is because the pre-trained CLIP text encoder takes limited number of textual tokens up to $77$, whereas the retrieval query can be long. For the cases where the tokenised text query is longer than the maximum supported length of CLIP, we simply truncate the sequence to fit our specified pattern, as such cases are few in practice.

\vspace{0.05cm}
When comparing with the Baseline-IV, which denotes the original CLIP using na\"ively-encoded text queries, our proposed prompt learning and temporal modeling clearly demonstrate benefits on all benchmarks. Additionally, while comparing to the existing approaches that are specifically targeting on retrieval, our method also performs competitively. Note that, we try to compare with the results reported in the existing work, however, this is by no means to be fair comparisons, as these methods are usually pre-trained on different datasets with variable sizes. For instance, CLIP4Clip~\cite{Luo_CLIP4Clip21} is pre-trained on HowTo100M~\cite{miech2019howto100m} (136M videos). Thus, our method is naturally at a disadvantage by only training on small-scale datasets. Moreover, in terms of computation cost, our method only optimises several prompt vectors, along with two Transformer layers, and all experiments can be done on \textit{one} 24G GeForce RTX 3090 GPU.

\section{Conclusion}
Building on CLIP, this paper constructs wide and strong baselines for efficient video understanding, with the simple idea of learning lightweight prompt vectors and temporal Transformer. We evaluate on $10$ popular benchmarks from: action recognition, action localisation, and text-video retrieval. Thorough comparisons and ablations are conducted to analyse the critical components. In the closed-set scenario, despite training only a few free parameters, we achieve competitive performance to the modern state-of-the-art methods. In few-shot and zero-shot scenarios, we significantly outperform existing methods on $7$ public benchmarks.

\vspace{0.15cm}
\noindent \textbf{Acknowledgements.} 
This work is supported by the National Key Research and Development Program of China (No. 2020YFB1406801), 111 plan (No. BP0719010), STCSM (No. 18DZ2270700), State Key Laboratory of UHD Video and Audio Production and Presentation, the UK EPSRC Programme Grant Visual AI (EP/T028572/1), and a Google-DeepMind Scholarship.

% \clearpage
\bibliographystyle{splncs04}
\bibliography{egbib}

\begin{thebibliography}{100}
\providecommand{\url}[1]{\texttt{#1}}
\providecommand{\urlprefix}{URL }
\providecommand{\doi}[1]{https://doi.org/#1}

\bibitem{anne2017localizing}
Anne~Hendricks, L., Wang, O., Shechtman, E., Sivic, J., Darrell, T., Russell,
  B.: Localizing moments in video with natural language. In: Proceedings of the
  International Conference on Computer Vision (2017)

\bibitem{arnab2021vivit}
Arnab, A., Dehghani, M., Heigold, G., Sun, C., Lu{\v{c}}i{\'c}, M., Schmid, C.:
  Vivit: A video vision transformer. In: Proceedings of the International
  Conference on Computer Vision (2021)

\bibitem{bain2021frozen}
Bain, M., Nagrani, A., Varol, G., Zisserman, A.: Frozen in time: A joint video
  and image encoder for end-to-end retrieval. Proceedings of the International
  Conference on Computer Vision  (2021)

\bibitem{Bertasius21}
Bertasius, G., Wang, H., Torresani, L.: Is space-time attention all you need
  for video understanding? In: Proceedings of the International Conference on
  Machine Learning (2021)

\bibitem{bishay2019tarn}
Bishay, M., Zoumpourlis, G., Patras, I.: Tarn: Temporal attentive relation
  network for few-shot and zero-shot action recognition. In: Proceedings of the
  British Machine Vision Conference (2019)

\bibitem{bodla2017soft}
Bodla, N., Singh, B., Chellappa, R., Davis, L.S.: Soft-nms--improving object
  detection with one line of code. In: Proceedings of the International
  Conference on Computer Vision (2017)

\bibitem{brattoli2020rethinking}
Brattoli, B., Tighe, J., Zhdanov, F., Perona, P., Chalupka, K.: Rethinking
  zero-shot video classification: End-to-end training for realistic
  applications. In: Proceedings of the IEEE Conference on Computer Vision and
  Pattern Recognition (2020)

\bibitem{Brown20}
Brown, T., Mann, B., Ryder, N., Subbiah, M., Kaplan, J.D., Dhariwal, P.,
  Neelakantan, A., Shyam, P., Sastry, G., Askell, A., Agarwal, S.,
  Herbert-Voss, A., Krueger, G., Henighan, T., Child, R., Ramesh, A., Ziegler,
  D., Wu, J., Winter, C., Hesse, C., Chen, M., Sigler, E., Litwin, M., Gray,
  S., Chess, B., Clark, J., Berner, C., McCandlish, S., Radford, A., Sutskever,
  I., Amodei, D.: Language models are few-shot learners. In: Advances in Neural
  Information Processing Systems (2020)

\bibitem{buch2019end}
Buch, S., Escorcia, V., Ghanem, B., Fei-Fei, L., Niebles, J.C.: End-to-end,
  single-stream temporal action detection in untrimmed videos. In: Proceedings
  of the British Machine Vision Conference (2019)

\bibitem{bulat2021space}
Bulat, A., Perez~Rua, J.M., Sudhakaran, S., Martinez, B., Tzimiropoulos, G.:
  Space-time mixing attention for video transformer. In: Advances in Neural
  Information Processing Systems (2021)

\bibitem{Cao20}
Cao, K., Ji, J., Cao, Z., Chang, C.Y., Niebles, J.C.: Few-shot video
  classification via temporal alignment. In: Proceedings of the IEEE Conference
  on Computer Vision and Pattern Recognition (2020)

\bibitem{Carreira19}
Carreira, J., Noland, E., Hillier, C., Zisserman, A.: A short note on the
  kinetics-700 human action dataset. arXiv preprint arXiv:1907.06987  (2019)

\bibitem{carreira2017quo}
Carreira, J., Zisserman, A.: Quo vadis, action recognition? a new model and the
  kinetics dataset. In: Proceedings of the IEEE Conference on Computer Vision
  and Pattern Recognition (2017)

\bibitem{Chao18}
Chao, Y.W., Vijayanarasimhan, S., Seybold, B., Ross, D.A., Deng, J.,
  Sukthankar, R.: Rethinking the faster r-cnn architecture for temporal action
  localisation. In: Proceedings of the IEEE Conference on Computer Vision and
  Pattern Recognition (2018)

\bibitem{croitoru2021teachtext}
Croitoru, I., Bogolin, S.V., Leordeanu, M., Jin, H., Zisserman, A., Albanie,
  S., Liu, Y.: Teachtext: Crossmodal generalized distillation for text-video
  retrieval. In: Proceedings of the International Conference on Computer Vision
  (2021)

\bibitem{Dwivedi19}
Dwivedi, S.K., Gupta, V., Mitra, R., Ahmed, S., Jain, A.: Protogan: Towards few
  shot learning for action recognition. In: Proceedings of the IEEE Conference
  on Computer Vision and Pattern Recognition (2019)

\bibitem{Dzabraev_2021_CVPR}
Dzabraev, M., Kalashnikov, M., Komkov, S., Petiushko, A.: Mdmmt: Multidomain
  multimodal transformer for video retrieval. In: Proceedings of the IEEE
  Conference on Computer Vision and Pattern Recognition Workshops (2021)

\bibitem{Feichtenhofer20}
Feichtenhofer, C.: {X3D: Expanding Architectures for Efficient Video
  Recognition}. In: Proceedings of the IEEE Conference on Computer Vision and
  Pattern Recognition (2020)

\bibitem{feichtenhofer2020x3d}
Feichtenhofer, C.: X3d: Expanding architectures for efficient video
  recognition. In: Proceedings of the IEEE Conference on Computer Vision and
  Pattern Recognition (2020)

\bibitem{Feichtenhofer19}
Feichtenhofer, C., Fan, H., Malik, J., He, K.: {SlowFast Networks for Video
  Recognition}. In: Proceedings of the International Conference on Computer
  Vision (2019)

\bibitem{Feichtenhofer16}
Feichtenhofer, C., Pinz, A., Zisserman, A.: Convolutional two-stream network
  fusion for video action recognition. In: Proceedings of the IEEE Conference
  on Computer Vision and Pattern Recognition (2016)

\bibitem{Frome13}
Frome, A., Corrado, G.S., Shlens, J., Bengio, S., Dean, J., Ranzato, M.A.,
  Mikolov, T.: Devise: A deep visual-semantic embedding model. In: Advances in
  Neural Information Processing Systems (2013)

\bibitem{gabeur2020multi}
Gabeur, V., Sun, C., Alahari, K., Schmid, C.: Multi-modal transformer for video
  retrieval. In: Proceedings of the European Conference on Computer Vision
  (2020)

\bibitem{Gabeur20}
Gabeur, V., Sun, C., Alahari, K., Schmid, C.: Multi-modal transformer for video
  retrieval. In: Proceedings of the European Conference on Computer Vision
  (2020)

\bibitem{Gan16a}
Gan, C., Yang, T., Gongi, B.: Learning attributes equals multi-source domain
  generalization. In: Proceedings of the IEEE Conference on Computer Vision and
  Pattern Recognition (2016)

\bibitem{Gan16b}
Gan, C., Yang, Y., Zhu, L., Zhao, D., Zhuang, Y.: Recognizing an action using
  its name: A knowledge-based approach. International Journal of Computer
  Vision  (2016)

\bibitem{gao2021clip}
Gao, P., Geng, S., Zhang, R., Ma, T., Fang, R., Zhang, Y., Li, H., Qiao, Y.:
  Clip-adapter: Better vision-language models with feature adapters. arXiv
  preprint arXiv:2110.04544  (2021)

\bibitem{Gao21}
Gao, T., Fisch, A., Chen, D.: Making pre-trained language models better
  few-shot learners. In: Association for Computational Linguistics (2021)

\bibitem{gao2019video}
Gao, Z., Wang, L., Zhang, Q., Niu, Z., Zheng, N., Hua, G.: Video imprint
  segmentation for temporal action detection in untrimmed videos. In:
  Proceedings of the AAAI Conference on Artificial Intelligence (2019)

\bibitem{goyal2017something}
Goyal, R., Ebrahimi~Kahou, S., Michalski, V., Materzynska, J., Westphal, S.,
  Kim, H., Haenel, V., Fruend, I., Yianilos, P., Mueller-Freitag, M., et~al.:
  The "something something" video database for learning and evaluating visual
  common sense. In: Proceedings of the International Conference on Computer
  Vision (2017)

\bibitem{Ha16}
Ha, D., Dai, A., Le, Q.: Hypernetworks. In: Proceedings of the International
  Conference on Learning Representations (2016)

\bibitem{han2022align}
Han, T., Xie, W., Zisserman, A.: Temporal alignment network for long-term
  video. In: Proceedings of the IEEE Conference on Computer Vision and Pattern
  Recognition (2022)

\bibitem{hara2018can}
Hara, K., Kataoka, H., Satoh, Y.: Can spatiotemporal 3d cnns retrace the
  history of 2d cnns and imagenet? In: Proceedings of the IEEE Conference on
  Computer Vision and Pattern Recognition (2018)

\bibitem{Caba15}
Heilbron, F.C., Escorcia, V., Ghanem, B., Niebles, J.C.: Activitynet: A
  large-scale video benchmark for human activity understanding. In: Proceedings
  of the IEEE Conference on Computer Vision and Pattern Recognition (2015)

\bibitem{jain2015objects2action}
Jain, M., Van~Gemert, J.C., Mensink, T., Snoek, C.G.: Objects2action:
  Classifying and localizing actions without any video example. In: Proceedings
  of the International Conference on Computer Vision (2015)

\bibitem{jain201515}
Jain, M., Van~Gemert, J.C., Snoek, C.G.: What do 15,000 object categories tell
  us about classifying and localizing actions? In: Proceedings of the IEEE
  Conference on Computer Vision and Pattern Recognition (2015)

\bibitem{Jia21}
Jia, C., Yang, Y., Xia, Y., Chen, Y.T., Parekh, Z., Pham, H., Le, Q.V., Sung,
  Y., Li, Z., Duerig, T.: Scaling up visual and vision-language representation
  learning with noisy text supervision. In: Proceedings of the International
  Conference on Machine Learning (2021)

\bibitem{jia2022visual}
Jia, M., Tang, L., Chen, B.C., Cardie, C., Belongie, S., Hariharan, B., Lim,
  S.N.: Visual prompt tuning. arXiv preprint arXiv:2203.12119  (2022)

\bibitem{THUMOS14}
Jiang, Y.G., Liu, J., Zamir, A.R., Toderici, G., Laptev, I., Shah, M.,
  Sukthankar, R.: {THUMOS} challenge: Action recognition with a large number of
  classes. \url{http://crcv.ucf.edu/THUMOS14/} (2014)

\bibitem{Jiang20}
Jiang, Z., Xu, F.F., Araki, J., Neubig, G.: How can we know what language
  models know? Transactions of the Association for Computational Linguistics
  (2020)

\bibitem{ju2021divide}
Ju, C., Zhao, P., Chen, S., Zhang, Y., Wang, Y., Tian, Q.: Divide and conquer
  for single-frame temporal action localization. In: Proceedings of the
  International Conference on Computer Vision (2021)

\bibitem{ju2021adaptive}
Ju, C., Zhao, P., Chen, S., Zhang, Y., Zhang, X., Tian, Q.: Adaptive mutual
  supervision for weakly-supervised temporal action localization. arXiv
  preprint arXiv:2104.02357  (2021)

\bibitem{kay2017}
Kay, W., Carreira, J., Simonyan, K., Zhang, B., Hillier, C., Vijayanarasimhan,
  S., Viola, F., Green, T., Back, T., Natsev, P., Suleyman, M., Zisserman, A.:
  The kinetics human action video dataset. arXiv preprint arXiv:1705.06950
  (2017)

\bibitem{Kuehne11}
Kuehne, H., Jhuang, H., Garrote, E., Poggio, T., Serre, T.: {HMDB}: A large
  video database for human motion recognition. In: Proceedings of the
  International Conference on Computer Vision (2011)

\bibitem{Lei21}
Lei, J., Li, L., Zhou, L., Gan, Z., Berg, T.L., Bansal, M., Liu, J.: Less is
  more: Clipbert for video-and-language learningvia sparse sampling. In:
  Proceedings of the IEEE Conference on Computer Vision and Pattern Recognition
  (2021)

\bibitem{Lester21}
Lester, B., Al-Rfou, R., Constant, N.: The power of scale for
  parameter-efficient prompt tuning. In: Proceedings of the Conference on
  Empirical Methods in Natural Language Processinng (2021)

\bibitem{li21-prefixtuning}
Li, X.L., Liang, P.: Prefix-tuning: Optimizing continuous prompts for
  generation. In: Association for Computational Linguistics (2021)

\bibitem{Li16}
Li, Y., hung Hu, S., Li, B.: Recognizing unseen actions in a domain-adapted
  embedding space. In: IEEE International Conference on Image Processing (2016)

\bibitem{lin2021learning}
Lin, C., Xu, C., Luo, D., Wang, Y., Tai, Y., Wang, C., Li, J., Huang, F., Fu,
  Y.: Learning salient boundary feature for anchor-free temporal action
  localization. In: Proceedings of the IEEE Conference on Computer Vision and
  Pattern Recognition (2021)

\bibitem{lin2019tsm}
Lin, J., Gan, C., Han, S.: Tsm: Temporal shift module for efficient video
  understanding. In: Proceedings of the International Conference on Computer
  Vision (2019)

\bibitem{lin2019bmn}
Lin, T., Liu, X., Li, X., Ding, E., Wen, S.: Bmn: Boundary-matching network for
  temporal action proposal generation. In: Proceedings of the International
  Conference on Computer Vision (2019)

\bibitem{lin2017single}
Lin, T., Zhao, X., Shou, Z.: Single shot temporal action detection. In:
  Proceedings of the ACM international conference on Multimedia (2017)

\bibitem{Lin18}
Lin, T., Zhao, X., Su, H., Wang, C., Yang, M.: {BSN}: Boundary sensitive
  network for temporal action proposal generation. In: Proceedings of the
  European Conference on Computer Vision (2018)

\bibitem{Liu11}
Liu, J., Kuipers, B., Savarese, S.: Recognizing human actions by attributes.
  In: Proceedings of the IEEE Conference on Computer Vision and Pattern
  Recognition (2011)

\bibitem{liu2019use}
Liu, Y., Albanie, S., Nagrani, A., Zisserman, A.: Use what you have: Video
  retrieval using representations from collaborative experts. In: Proceedings
  of the British Machine Vision Conference (2019)

\bibitem{liu2022video}
Liu, Z., Ning, J., Cao, Y., Wei, Y., Zhang, Z., Lin, S., Hu, H.: Video swin
  transformer. In: Proceedings of the IEEE Conference on Computer Vision and
  Pattern Recognition (2022)

\bibitem{loshchilov2017decoupled}
Loshchilov, I., Hutter, F.: Decoupled weight decay regularization. In:
  Proceedings of the International Conference on Learning Representations
  (2019)

\bibitem{Luo_CLIP4Clip21}
Luo, H., Ji, L., Zhong, M., Chen, Y., Lei, W., Duan, N., Li, T.: {CLIP4Clip}:
  An empirical study of clip for end to end video clip retrieval. arXiv
  preprint arXiv:2104.08860  (2021)

\bibitem{Mettes21}
Mettes, P., Thong, W., Snoek, C.G.M.: Object priors for classifying and
  localizing unseen actions. In: International Journal of Computer Vision
  (2021)

\bibitem{Miech20}
Miech, A., Alayrac, J.B., Smaira, L., Laptev, I., Sivic, J., Zisserman, A.:
  End-to-end learning of visual representations from uncurated instructional
  videos. In: Proceedings of the IEEE Conference on Computer Vision and Pattern
  Recognition (2020)

\bibitem{Miech18}
Miech, A., Laptev, I., Sivic, J.: Learning a text-video embedding from
  imcomplete and heterogeneous data. arXiv preprint arXiv:1804.02516  (2018)

\bibitem{miech2019howto100m}
Miech, A., Zhukov, D., Alayrac, J.B., Tapaswi, M., Laptev, I., Sivic, J.:
  Howto100m: Learning a text-video embedding by watching hundred million
  narrated video clips. In: Proceedings of the International Conference on
  Computer Vision (2019)

\bibitem{Mishra20}
Mishra, A., Pandey, A., Murthy, H.A.: Zero-shot learning for action recognition
  using synthesized features. Neurocomputing  (2020)

\bibitem{monfort2021spoken}
Monfort, M., Jin, S., Liu, A., Harwath, D., Feris, R., Glass, J., Oliva, A.:
  Spoken moments: Learning joint audio-visual representations from video
  descriptions. In: Proceedings of the IEEE Conference on Computer Vision and
  Pattern Recognition (2021)

\bibitem{monfort2021multi}
Monfort, M., Pan, B., Ramakrishnan, K., Andonian, A., McNamara, B.A.,
  Lascelles, A., Fan, Q., Gutfreund, D., Feris, R., Oliva, A.: Multi-moments in
  time: Learning and interpreting models for multi-action video understanding.
  IEEE Transactions on Pattern Analysis and Machine Intelligence  (2021)

\bibitem{Mori99}
Mori, Y., Takahashi, H., Oka, R.: Image-to-word transformation based on
  dividing and vector quantizing images with words. In: First International
  Workshop on Multimedia Intelligent Storage and Retrieval Management (ACM
  Multimedia Conference) (1999)

\bibitem{Nawhal21}
Nawhal, M., Mori, G.: Activity graph transformer for temporal action
  localization. arXiv preprint arXiv:2101.08540  (2021)

\bibitem{Perrett21}
Perrett, T., Masullo, A., Burghardt, T., Mirmehdi, M., Damen, D.: Temporal
  relational crosstransformers for few-shot action recognition. In: Proceedings
  of the IEEE Conference on Computer Vision and Pattern Recognition (2021)

\bibitem{Radford21}
Radford, A., Kim, J.W., Hallacy, C., Ramesh, A., Goh, G., Agarwal, S., Sastry,
  G., Askell, A., Mishkin, P., Clark, J., Krueger, G., Sutskever, I.: Learning
  transferable visual models from natural language supervision. In: Proceedings
  of the International Conference on Machine Learning (2021)

\bibitem{rohrbach2017movie}
Rohrbach, A., Torabi, A., Rohrbach, M., Tandon, N., Pal, C., Larochelle, H.,
  Courville, A., Schiele, B.: Movie description. International Journal of
  Computer Vision  (2017)

\bibitem{Timo21}
Schick, T., Schütze, H.: Exploiting cloze questions for few shot text
  classification and natural language inference. In: Proceedings of the
  Conference of the European Chapter of the Association for Computational
  Linguistics (2021)

\bibitem{Shin20}
Shin, T., Razeghi, Y., IV, R.L.L., Wallace, E., Singh, S.: {AutoPrompt}:
  Eliciting knowledge from language models with automatically generated
  prompts. In: Proceedings of the Conference on Empirical Methods in Natural
  Language Processinng (2020)

\bibitem{shou2017cdc}
Shou, Z., Chan, J., Zareian, A., Miyazawa, K., Chang, S.F.: Cdc:
  Convolutional-de-convolutional networks for precise temporal action
  localization in untrimmed videos. In: Proceedings of the IEEE Conference on
  Computer Vision and Pattern Recognition (2017)

\bibitem{Shou16}
Shou, Z., Wang, D., Chang, S.F.: Temporal action localization in untrimmed
  videos via multi-stage cnns. In: Proceedings of the IEEE Conference on
  Computer Vision and Pattern Recognition (2016)

\bibitem{Simonyan14}
Simonyan, K., Zisserman, A.: Two-stream convolutional networks for action
  recognition in videos. In: Advances in Neural Information Processing Systems
  (2014)

\bibitem{Soomro12}
Soomro, K., Zamir, A.R., Shah, M.: {UCF101}: A dataset of 101 human actions
  classes from videos in the wild. arXiv preprint arXiv:1212.0402  (2012)

\bibitem{su2020bsn++}
Su, H., Gan, W., Wu, W., Qiao, Y., Yan, J.: {Bsn++}: Complementary boundary
  regressor with scale-balanced relation modeling for temporal action proposal
  generation. In: Proceedings of the AAAI Conference on Artificial Intelligence
  (2021)

\bibitem{tan2021relaxed}
Tan, J., Tang, J., Wang, L., Wu, G.: Relaxed transformer decoders for direct
  action proposal generation. In: Proceedings of the International Conference
  on Computer Vision (2021)

\bibitem{Tran18}
Tran, D., Wang, H., Torresani, L., Ray, J., LeCun, Y., Paluri, M.: A closer
  look at spatiotemporal convolutions for action recognition. In: Proceedings
  of the IEEE Conference on Computer Vision and Pattern Recognition (2018)

\bibitem{Wang16-TSN}
Wang, L., Xiong, Y., Wang, Z., Qiao, Y., Lin, D., Tang, X., {Val Gool}, L.:
  Temporal segment networks: Towards good practices for deep action
  recognition. In: Proceedings of the European Conference on Computer Vision
  (2016)

\bibitem{Wang_ActionCLIP21}
Wang, M., Xing, J., Liu, Y.: {ActionCLIP}: A new paradigm for video action
  recognition. arXiv preprint arXiv:2109.08472  (2021)

\bibitem{wang2022rcl}
Wang, Q., Zhang, Y., Zheng, Y., Pan, P.: Rcl: Recurrent continuous localization
  for temporal action detection. In: Proceedings of the IEEE Conference on
  Computer Vision and Pattern Recognition (2022)

\bibitem{wang2018non}
Wang, X., Girshick, R., Gupta, A., He, K.: Non-local neural networks. In:
  Proceedings of the IEEE Conference on Computer Vision and Pattern Recognition
  (2018)

\bibitem{Weston11}
Weston, J., Bengio, S., Usunier, N.: {WSABIE}: Scaling up to large vocabulary
  image annotation. In: Proceedings of the International Joint Conference on
  Artificial Intelligence (2011)

\bibitem{Xie18-S3D}
Xie, S., Sun, C., Huang, J., Tu, Z., Murphy, K.: Rethinking spatiotemporal
  feature learning for video understanding. In: Proceedings of the European
  Conference on Computer Vision (2018)

\bibitem{Xu17}
Xu, H., Das, A., Saenko, K.: {R-C3D}: Region convolutional 3d network for
  temporal activity detection. In: Proceedings of the International Conference
  on Computer Vision (2017)

\bibitem{xu2016msr}
Xu, J., Mei, T., Yao, T., Rui, Y.: {MSR-VTT}: A large video description dataset
  for bridging video and language. In: Proceedings of the IEEE Conference on
  Computer Vision and Pattern Recognition (2016)

\bibitem{xu2020g}
Xu, M., Zhao, C., Rojas, D.S., Thabet, A., Ghanem, B.: G-tad: Sub-graph
  localization for temporal action detection. In: Proceedings of the IEEE
  Conference on Computer Vision and Pattern Recognition (2020)

\bibitem{yang2020revisiting}
Yang, L., Peng, H., Zhang, D., Fu, J., Han, J.: Revisiting anchor mechanisms
  for temporal action localization. IEEE Transactions on Image Processing
  (2020)

\bibitem{yao2021filip}
Yao, L., Huang, R., Hou, L., Lu, G., Niu, M., Xu, H., Liang, X., Li, Z., Jiang,
  X., Xu, C.: Filip: Fine-grained interactive language-image pre-training. In:
  Proceedings of the International Conference on Learning Representations
  (2022)

\bibitem{Yeung16}
Yeung, S., Russakovsky, O., Mori, G., Fei-Fei, L.: End-to-end learning of
  action detection from frame glimpses in videos. In: Proceedings of the IEEE
  Conference on Computer Vision and Pattern Recognition (2016)

\bibitem{yu2018joint}
Yu, Y., Kim, J., Kim, G.: A joint sequence fusion model for video question
  answering and retrieval. In: Proceedings of the European Conference on
  Computer Vision (2018)

\bibitem{zhang2022actionformer}
Zhang, C., Wu, J., Li, Y.: Actionformer: Localizing moments of actions with
  transformers. arXiv preprint arXiv:2202.07925  (2022)

\bibitem{Zhang20a}
Zhang, H., Zhang, L., Qi, X., Li, H., Torr, P.H.S., Koniusz, P.: Few-shot
  action recognition with permutation-invariant attention. In: Proceedings of
  the European Conference on Computer Vision (2020)

\bibitem{zhang2021tip}
Zhang, R., Fang, R., Gao, P., Zhang, W., Li, K., Dai, J., Qiao, Y., Li, H.:
  Tip-adapter: Training-free clip-adapter for better vision-language modeling.
  arXiv preprint arXiv:2111.03930  (2021)

\bibitem{zhao2020bottom}
Zhao, P., Xie, L., Ju, C., Zhang, Y., Wang, Y., Tian, Q.: Bottom-up temporal
  action localization with mutual regularization. In: Proceedings of the
  European Conference on Computer Vision (2020)

\bibitem{Zhao17}
Zhao, Y., Xiong, Y., Wang, L., Wu, Z., Tang, X., Lin, D.: Temporal action
  detection with structured segment networks. In: Proceedings of the
  International Conference on Computer Vision (2017)

\bibitem{zhou2018temporal}
Zhou, B., Andonian, A., Oliva, A., Torralba, A.: Temporal relational reasoning
  in videos. In: Proceedings of the European Conference on Computer Vision
  (2018)

\bibitem{zhou2021learning}
Zhou, K., Yang, J., Loy, C.C., Liu, Z.: Learning to prompt for vision-language
  models. arXiv preprint arXiv:2109.01134  (2021)

\bibitem{zhou2022conditional}
Zhou, K., Yang, J., Loy, C.C., Liu, Z.: Conditional prompt learning for
  vision-language models. In: Proceedings of the IEEE Conference on Computer
  Vision and Pattern Recognition (2022)

\bibitem{Zhu18}
Zhu, L., Yang, Y.: Compound memory networks for few-shot video classification.
  In: Proceedings of the European Conference on Computer Vision (2018)

\bibitem{zhu2020label}
Zhu, L., Yang, Y.: Label independent memory for semi-supervised few-shot video
  classification. IEEE Transactions on Pattern Analysis and Machine
  Intelligence  (2020)

\bibitem{zhu2021few}
Zhu, X., Toisoul, A., Perez-Rua, J.M., Zhang, L., Martinez, B., Xiang, T.:
  Few-shot action recognition with prototype-centered attentive learning. In:
  Proceedings of the British Machine Vision Conference (2021)

\end{thebibliography}
\clearpage

\appendix

\section{Implementation Details}   \label{section:supp_details}
The image and text encoders are adopted from the pre-trained CLIP (ViT-B/16), and are both kept frozen. All prompt vectors and visual features are of the same dimension, $D = 512$, and the temperature hyper-parameter $\tau$ is set to $0.07$. Both prompt vectors and temporal Transformer are randomly initialised by drawing from a zero-mean Gaussian distribution with the standard deviation of 0.01. For action recognition and action localisation~(the second-stage proposal classifier), we evaluate different numbers of prompt vectors, and adopt the $[16 + 16]$ pattern eventually, {\em i.e.} $16$ random vectors are prepended / appended to the input text tokens, and optimised for the considered tasks. For text-video retrieval, as the text description can be long, we utilise $[4 + 4]$ prompt vectors. In terms of spatial pre-processing, we resize the frame's short side to $224$, while keeping its original aspect ratio, then perform center cropping to convert the spatial size to $224\times224$.

\subsection{Action Recognition}
For action recognition (and text-video retrieval), at inference time, we random sample $16$ frames from each video for $5$ times, and take the average of these $5$ results as the final predictions, {\em i.e.}~$5$-crop evaluation.

\subsection{Action Localisation}
For action localisation, to obtain class-agnostic action proposals, we adopt the off-the-shelf proposal detectors~\cite{lin2021learning,yang2020revisiting}. 

To be specific, we first divide the entire video into several equal-frame segments; use the CLIP image encoder with one Transformer layer to extract frame-wise embeddings; feed these embeddings to the $6$-layer feature pyramid; utilise three parallel prediction heads to determine the actionness, centerness, boundaries respectively; finally, assemble all prediction results and use Soft-NMS~\cite{bodla2017soft} to suppress redundant proposals. On ActivityNet1.3, we maintain the original video frame rate, and use $768$ frames in each segment. On THUMOS14, we downsample each video to $10$ fps, and $256$ frames are used to construct the segment. The proposal detector is optimised using AdamW~\cite{loshchilov2017decoupled} with a learning rate of $10^{-4}$, and a batch size of $32$ videos. Please refer to~\cite{lin2021learning,yang2020revisiting} for detailed architectures and optimisation objectives. For post-processing, we set the tIoU threshold in Soft-NMS to $0.5$ on THUMOS14, and $0.85$ on ActivityNet1.3.

\subsection{Text-Video Retrieval}
For text-video retrieval, all the videos are decoded with $30$ fps in advance, and we take the $16$-frame input with a random frame $\text{gap} \in \{10, 15, 30\}$, that is, the video is equivalent to being sampled with $1$-$3$ fps. Note that, here we adopt significantly lower fps than action recognition, as the video retrieval task tends to require information from long-term visual dependencies.

\section{Experiment and Analysis}
In this section, we demonstrate more results to further analyse our method, and explore the semantic information learnt by prompt vectors.

\subsection{Action Localisation}   \label{subsection:supp_openset_localisation}
We adopt the two-stage paradigm for action localisation, {\em i.e.}~first-stage proposal detection and second-stage proposal classification. In this section, we separately evaluate the performance of these two stages in closed-set and zero-shot scenarios, to comprehensively dissect localisation results.

% ==================================================
\begin{table*}[t]
\scriptsize
\caption{\footnotesize \textbf{Results of proposal detection.} For closed-set, we train and evaluate on the same action categories. While for zero-shot, we experiment with two settings: training with 75\% (25\%) categories and testing on the remaining 25\% (50\%) categories.}
\centering
\begin{tabular}{C{1.8cm}C{2.2cm}|C{1.6cm}C{1.6cm}|C{2.2cm}}
\toprule
\multirow{2}{*}{} & \multirow{2}{*}{} & \multicolumn{2}{c}{THUMOS14} & \multicolumn{1}{c}{ActivityNet1.3} \\ \midrule
Scenario & Train {\em v.s}\, Test & AR@50 & AR@100 & AR@100 \\ \midrule
\multicolumn{1}{c|}{Closed-set} & 100\% {\em v.s}\, 100\% & 32.4 & 38.3  & 63.6 \\ \midrule
\multicolumn{1}{c|}{Zero-shot} & 75\% {\em v.s}\, 25\% & 24.1 & 29.7 & 60.8 \\
\multicolumn{1}{c|}{Zero-shot} & 50\% {\em v.s}\, 50\% & 21.2 & 26.2 & 59.3 \\
\bottomrule
\end{tabular}
\label{tab:proposal detection}
\end{table*}
% ==================================================

% ==================================================
\begin{table*}[t]
\scriptsize
\caption{\footnotesize \textbf{Results of proposal classification.} For closed-set of THUMOS14 (ActivityNet1.3), we train and test on the same 20 (200) action categories. While for zero-shot, we experiment with two settings, training with 75\% (50\%) action categories and testing on the remaining 25\% (50\%) action categories, {\em e.g.} training on 15 (10) categories and evaluating on the left 5 (10) categories for THUMOS14.}
\centering
\begin{tabular}{C{1.8cm}C{2.2cm}|C{1.5cm}C{1.2cm}|C{1.5cm}C{1.2cm}}
\toprule
\multirow{2}{*}{} & \multirow{2}{*}{} & \multicolumn{2}{c}{THUMOS14} & \multicolumn{2}{c}{ActivityNet1.3} \\ \midrule
Scenario & Train {\em v.s}\, Test & train\,/\,test & TOP1 & train\,/\,test & TOP1 \\ \midrule
\multicolumn{1}{c|}{Closed-set} & 100\% {\em v.s}\, 100\%  & 20\,/\,20 & 88.7 & 200\,/\,200 & 85.6 \\ \midrule
\multicolumn{1}{c|}{Zero-shot} & 75\% {\em v.s}\, 25\%  & 15\,/\,5 & 93.4 & 150\,/\,50 & 81.5\\
\multicolumn{1}{c|}{Zero-shot} & 50\% {\em v.s}\, 50\% & 10\,/\,10 & 87.3 & 100\,/\,100 & 71.8 \\
\bottomrule
\end{tabular}
\label{tab:proposal classification}
\end{table*}
% ==================================================

\subsubsection{Proposal Detection.}
\hspace{5pt} To evaluate class-agnostic action proposals, we adopt conventional metric: Average Recall with different Average Number (AR@AN). On THUMOS14, the AR is calculated under multiple IoU threshold set from 0.5 to 1.0 with a stride of 0.05. As for ActivityNet1.3, the multiple IoU threshold are from 0.5 to 0.95 with a stride of 0.05. And for the zero-shot settings with multiple sampling trials, we average the AR of all trials.

\vspace{0.05cm}
Table~\ref{tab:proposal detection} shows the comparison results. On both datasets, the performance of the zero-shot scenario decreases compared with that of the closed-set scenario, showing that the action proposal is in fact not perfectly class-agnostic, it is still biased towards seen action categories. Moreover, since each video on THUMOS14 contains denser action instances, the number of which is $10$ times than that of ActivityNet1.3, the performance drop on THUMOS14 is more significant.

\subsubsection{Proposal Classification.}
\hspace{3pt} We eliminate the action proposals that are completely disjoint with all ground-truth action instances, and evaluate the standard TOP1 classification accuracy among the remaining action proposals.

\vspace{0.05cm}
Table~\ref{tab:proposal classification} shows the average accuracy of multiple sampling trials. Comparing to the closed-set evaluation, the zero-shot classification accuracy tends to drop. Note that, the setting training with 75\% action categories on THUMOS14 is a special case. Since THUMOS14 has total $20$ action categories, in this case, the number of testing categories is only $5$, thus the classification task is definitely easier than the closed-set scenario.

\subsubsection{Summary.}
The above results show that, the performance drop of the zero-shot scenario comes from two sources: one is the recall drop from the first-stage action proposals, and the other comes from the second-stage classification errors.

% ==================================================
\begin{table*}[t]
\scriptsize
\caption{\footnotesize \textbf{Results of text-video retrieval.} Baseline-IV refers to original CLIP with text query na\"ively encoded, {\em i.e.}~without using any prompt. \textsc{E2E} denotes if the model is trained end-to-end. We highlight the results without end-to-end finetuning, where the best and second-best results are highlighted with bold and underline. As these methods are pre-trained on different datasets with variable sizes, it is unlikely to compare fairly.}
\centering
\begin{tabular}{C{1.7cm}C{0.5cm}|cccc|cccc|cc|cc}
\toprule
\multirow{2}{*}{} & \multirow{2}{*}{} & \multicolumn{4}{c}{MSRVTT~(9K)} & \multicolumn{4}{c}{LSMDC} & \multicolumn{2}{c}{SMIT}  & \multicolumn{2}{c}{DiDeMo} \\ \midrule
Method & E2E & R@1$\uparrow$ & R@5$\uparrow$ & R@10$\uparrow$ & MdR$\downarrow$  & R@1$\uparrow$ & R@5$\uparrow$ & R@10$\uparrow$ & MdR$\downarrow$    & R@1$\uparrow$ & R@5$\uparrow$ & R@1$\uparrow$ & R@5$\uparrow$ \\ \midrule
CE~\cite{liu2019use} & \xmark & 21.7 & 51.8  & 65.7 & 5.0 & 12.4 & 28.5  & 37.9& 21.7& -- & -- & 16.1 & 41.1 \\ 
MMT~\cite{gabeur2020multi} & \xmark & 24.6 & 54.0 & 67.1 & 4.0 & 13.2 & 29.2 & 38.8 & 21.0 & -- & -- & -- & -- \\ 
TT-CE+~\cite{croitoru2021teachtext} & \xmark & 29.6 & 61.6  & 74.2 & {3.0} & \underline{17.2} & \underline{36.5}  & \underline{46.3} & \underline{13.7} & -- & -- & 21.6 & 48.6 \\ 
SMiT~\cite{monfort2021spoken} & \xmark & 33.1 & \underline{64.8}  & \underline{77.4} & -- & -- & --  & -- & -- & \underline{39.5} & \underline{65.7} & -- & --\\
MDMMT~\cite{Dzabraev_2021_CVPR} & \xmark & {\bf 38.9} & {\bf 69.0}  & {\bf 79.7} & {\bf 2.0} & {\bf 18.8} & {\bf 38.5}  & {\bf 47.9} & {\bf 12.3} & --& -- & -- & --\\
Baseline-IV & \xmark & 31.2 & 53.7 & 64.2 & 4.0  & 11.3 & 22.7 & 29.2 & 56.5  & 39.3 & 62.8 & \underline{28.8} & \underline{54.6} \\ 
Ours & \xmark & {\underline{36.7}} & {64.6} & {76.8}  & {\bf 2.0} & {13.4} & {29.5} & {40.3}  & {18.6}  &  {\bf 66.6} & {\bf 87.8}  & \textbf{36.1} & \textbf{64.8}\\ \midrule
Frozen~\cite{bain2021frozen} & \checkmark & 31.0 & 59.5 & 70.5 & 3.0 & 15.0 & 30.8 & 39.8 & 20.0&  -- & -- & 34.6 & 65.0 \\ 
CLIP4Clip~\cite{Luo_CLIP4Clip21} & \checkmark & 44.5 & 71.4 & 81.6 & 2.0 & 22.6 & 41.0  & 49.1 & 11.0 & -- & -- & 43.4 & 70.2 \\  \bottomrule
\end{tabular}
\label{tab:text-video retrieval}
\end{table*}
% ==================================================

\subsection{Text-Video Retrieval}
Here, we add more comparison results for retrieval benchmarks. Since the text encoder from the pre-trained CLIP takes limited number of textual tokens up to $77$, whereas the text query of retrieval can be long, in these experiments, we only employ $8$ learnable prompt vectors, {\em i.e.}~[4+X+4]. And for temporal modeling, we only use two Transformer layers to achieve efficient model adaptation.

As can be observed in Table~\ref{tab:text-video retrieval}, our method with learnable prompt vectors largely outperforms the Baseline-IV, which refers to the original CLIP with text query na\"ively encoded, {\em i.e.}~without using any prompt. Despite only optimising a few parameters, our performance are sometimes comparable to the state-of-the-art methods that have been specifically designed for retrieval.

\subsection{Prompt Semantics}
To demonstrate the prompt semantics, we visualise the learnt $32$ prompt vectors, by searching for the word embeddings whose cosine distance is nearest to them. Here, we regard the CLIP vocabulary library as the total search set, {\em i.e.} 49408 subwords. For HMDB51 under the closed-set scenario, the nearest subwords are ``\textit{educ, Ā, giggle, meyers, lucas, windows, resolution, fives, me, lump, chancellor, extensively, previously, trades, sden, bowler, giuliani, radi, ivory, ffey, plays, evolu, acies, ghead, forsyth, botanic, unite, \&, protestant, saucer, ferry, mango}''.

\vspace{0.05cm}
As can be seen, some searched subwords are related to datasets or tasks, but most do not correspond to meaningful semantics. Such phenomenon about learnt prompt semantics, is in accordance with the observation in the NLP domain~\cite{Lester21}. We speculate this is because, the prompt vectors learnt in continuous embedding space go beyond discrete vocabulary space. In other words, the CLIP vocabulary library is limited to interpret the learnt prompt semantics.

\section{Limitations}
Our proposed idea relies on the visual-language model pre-trained on the large-scale image alt-text data, which could potentially incur two limitations: {\em First}, bias in the web data. {\em Second}, as temporal modeling is only used on top of visual features, it may fail to model fine-grained motions.

\section{Dataset Splits}
Here, we detail the dataset splits for training and testing, under different scenarios, namely, few-shot action recognition, zero-shot action recognition, and zero-shot action localisation. All the splits can be available at https://github.com/ju-chen/Efficient-Prompt/tree/main/datasplits.

\subsection{Few-shot Action Recognition}  \label{subsection:datasplit_fewshot_recognition}
\subsubsection{$5$-Shot-$5$-Way Setting.} 
\hspace{3pt} We here adopt the publicly available few-shot data splits, {\em i.e.}~sample $5$ action categories ($5$ videos per category) from a set of testing categories, to form the few-shot support set. We conduct $200$ trials with random samplings, to ensure the statistical significance.

\begin{itemize}
\item {\bf Kinetics-400}. 
We follow~\cite{Zhu18,Perrett21} and sample the test action categories from: blasting sand, busking, cutting watermelon, dancing ballet, dancing charleston, dancing macarena, diving cliff, filling eyebrows, folding paper, hula hooping, hurling (sport), ice skating, paragliding, playing drums, playing monopoly, playing trumpet, pushing car, riding elephant, shearing sheep, side kick, stretching arm, tap dancing, throwing axe, unboxing.

\item {\bf UCF-101}. 
Following~\cite{Zhang20a}, the test action categories are sampled from : blowingcandles, cleanandjerk, cliffdiving, cuttinginkitchen, diving, floorgymnastics, golfswing, handstandwalking, horserace, icedancing, jumprope, pommelhorse, punch, rockclimbingindoor, salsaspin, skiing, skydiving, stillrings, surfing, tennisswing, volleyballspiking.

\item {\bf HMDB-51}. 
We follow~\cite{Zhang20a} and sample the test action categories from: fencing, kick, kick ball, pick, pour, pushup, run, sit, smoke, talk.
\end{itemize}

\subsubsection{$5$-Shot-$C$-Way Setting.}
\hspace{3pt} In this generalised problem, to construct the dataset for training, we sample $5$ videos from all categories and measure the performance on the standard testing set, {\em i.e.}~all videos from all categories in the testing set. We also conduct $10$ random sampling rounds to choose training videos.

\begin{itemize}
    \item {\bf Kinetics-400}. Its training set contains $2000$ videos, {\em i.e.}~$400 \times 5$ videos, and the testing set covers $19101$ videos.
    \item {\bf UCF-101}. The training set contains $505$ videos, {\em i.e.}~$101 \times 5$ videos, and the testing set covers $3783$ videos.
    \item {\bf HMDB-51}. The training data covers $255$ videos, {\em i.e.}~$51 \times 5$ videos, and the testing set contains $1530$ videos.
\end{itemize}

\subsection{Zero-shot Action Recognition} 
In this section, we split K-700 dataset into two subsets with disjoint categories. Specifically, $400$ action categories are utilised for training, and the remaining $300$ action categories are used for evaluation.

\begin{itemize}
\item {\bf Training Categories~(\#400):}
carving wood with a knife,
cracking neck,
feeding goats,
fixing bicycle,
passing soccer ball,
being in zero gravity,
breaking boards,
changing gear in car,
playing organ,
taking photo,
finger snapping,
walking on stilts,
cleaning shoes,
hoverboarding,
putting wallpaper on wall,
using atm,
rock scissors paper,
riding elephant,
running on treadmill,
cracking back,
pulling rope (game),
washing feet,
skydiving,
country line dancing,
throwing knife,
square dancing,
fixing hair,
folding clothes,
doing jigsaw puzzle,
making slime,
using a power drill,
welding,
jumping jacks,
cosplaying,
surveying,
bottling,
smoking pipe,
shooting basketball,
swimming with dolphins,
tying bow tie,
cleaning gutters,
playing cards,
playing dominoes,
uncorking champagne,
drop kicking,
folding paper,
standing on hands,
massaging neck,
swing dancing,
chopping meat,
breading or breadcrumbing,
laying concrete,
driving car,
sawing wood,
clean and jerk,
embroidering,
pinching,
playing saxophone,
tango dancing,
peeling banana,
drumming fingers,
throwing axe,
lawn mower racing,
roller skating,
celebrating,
dyeing eyebrows,
arm wrestling,
belly dancing,
using segway,
playing cello,
news anchoring,
mountain climber (exercise),
treating wood,
riding mechanical bull,
cutting watermelon,
playing laser tag,
picking apples,
using a sledge hammer,
skipping rope,
feeding fish,
playing basketball,
carving pumpkin,
bee keeping,
holding snake,
walking through snow,
fly tying,
tightrope walking,
playing monopoly,
shopping,
planing wood,
brushing floor,
cleaning pool,
spinning poi,
grooming horse,
laughing,
sign language interpreting,
roasting pig,
making cheese,
ripping paper,
decorating the christmas tree,
spraying,
snowkiting,
putting on shoes,
playing cricket,
ironing,
mosh pit dancing,
swimming butterfly stroke,
ironing hair,
making the bed,
chiseling stone,
javelin throw,
playing keyboard,
poaching eggs,
playing recorder,
blowing nose,
high kick,
shot put,
tasting beer,
laying tiles,
making paper aeroplanes,
being excited,
parkour,
playing piano,
throwing discus,
wading through mud,
washing dishes,
headbutting,
tying knot (not on a tie),
unloading truck,
visiting the zoo,
picking blueberries,
gymnastics tumbling,
playing checkers,
hugging baby,
playing netball,
spray painting,
attending conference,
playing trombone,
using bagging machine,
listening with headphones,
making sushi,
trimming or shaving beard,
swimming with sharks,
throwing water balloon,
plastering,
playing pan pipes,
directing traffic,
assembling computer,
making horseshoes,
ice swimming,
pull ups,
battle rope training,
blowdrying hair,
doing laundry,
ice skating,
shouting,
surfing water,
barbequing,
vacuuming floor,
squat,
dribbling basketball,
chasing,
throwing ball (not baseball or American football),
eating doughnuts,
contact juggling,
deadlifting,
dancing gangnam style,
pretending to be a statue,
shaving head,
putting on eyeliner,
blowing bubble gum,
jumping into pool,
juggling fire,
grinding meat,
moving furniture,
tagging graffiti,
skiing mono,
bookbinding,
walking the dog,
petting animal (not cat),
falling off bike,
scrambling eggs,
sipping cup,
separating eggs,
historical reenactment,
springboard diving,
eating watermelon,
card throwing,
using a microscope,
playing poker,
making pizza,
assembling bicycle,
backflip (human),
seasoning food,
getting a tattoo,
shining shoes,
snatch weight lifting,
installing carpet,
getting a haircut,
laying decking,
rock climbing,
sieving,
rope pushdown,
opening bottle (not wine),
salsa dancing,
catching or throwing baseball,
texting,
clapping,
mopping floor,
pirouetting,
scuba diving,
coughing,
climbing a rope,
changing oil,
yarn spinning,
playing guitar,
using a paint roller,
snowmobiling,
tying necktie,
vacuuming car,
petting horse,
busking,
paragliding,
playing kickball,
chewing gum,
giving or receiving award,
drooling,
putting in contact lenses,
alligator wrestling,
doing aerobics,
whistling,
somersaulting,
carrying baby,
decoupage,
slicing onion,
jetskiing,
carving ice,
baking cookies,
checking watch,
rolling pastry,
pumping fist,
crocheting,
eating burger,
jumping sofa,
dodgeball,
karaoke,
waxing back,
leatherworking,
passing American football (not in game),
massaging feet,
dumpster diving,
making balloon shapes,
cracking knuckles,
eating spaghetti,
catching or throwing frisbee,
drinking shots,
playing gong,
acting in play,
shoveling snow,
sharpening knives,
using megaphone,
doing nails,
burping,
inflating balloons,
flying kite,
herding cattle,
doing sudoku,
eating hotdog,
putting on sari,
punching bag,
singing,
squeezing orange,
pushing cart,
splashing water,
playing trumpet,
exercising arm,
fencing (sport),
ski jumping,
lock picking,
carrying weight,
using inhaler,
waking up,
staring,
photobombing,
eating carrots,
bungee jumping,
checking tires,
weaving fabric,
home roasting coffee,
playing didgeridoo,
getting a piercing,
building cabinet,
jumping bicycle,
capoeira,
reading newspaper,
playing rubiks cube,
high jump,
raising eyebrows,
stretching arm,
shooting off fireworks,
dancing charleston,
pillow fight,
hockey stop,
steering car,
drawing,
recording music,
front raises,
riding camel,
wrapping present,
waxing legs,
sleeping,
cooking scallops,
sucking lolly,
cutting cake,
threading needle,
base jumping,
dining,
trapezing,
tackling,
building shed,
tiptoeing,
cooking chicken,
playing harmonica,
training dog,
setting table,
curling eyelashes,
passing American football (in game),
docking boat,
playing paintball,
sneezing,
playing with trains,
swimming breast stroke,
sticking tongue out,
cutting pineapple,
lunge,
triple jump,
marriage proposal,
cleaning windows,
diving cliff,
bench pressing,
making a cake,
saluting,
luge,
driving tractor,
swimming front crawl,
bending back,
laying stone,
pushing car,
sanding wood,
dunking basketball,
sanding floor,
sausage making,
robot dancing,
building sandcastle,
tasting food,
spelunking,
baby waking up,
playing darts,
playing american football,
land sailing,
sword fighting,
ski ballet,
playing mahjong,
smelling feet,
blasting sand,
peeling potatoes,
smoking,
hurdling,
grooming cat,
pouring beer,
bobsledding,
flint knapping,
washing hands,
clay pottery making,
digging,
air drumming,
moving child,
fidgeting,
packing,
delivering mail,
skipping stone,
cartwheeling,
playing bass guitar,
tai chi,
using remote controller (not gaming),
playing pinball,
bartending,
waxing chest,
parasailing,
egg hunting,
carving marble,
wrestling,
snowboarding,
headbanging,
playing hand clapping games,
abseiling,
crawling baby,
skiing slalom,
frying vegetables,
wading through water.
\item {\bf Testing Categories~(\#300):} 
adjusting glasses,
answering questions,
applauding,
applying cream,
archaeological excavation,
archery,
arguing,
arranging flowers,
arresting,
auctioning,
bandaging,
bathing dog,
beatboxing,
bending metal,
biking through snow,
blending fruit,
blowing glass,
blowing leaves,
blowing out candles,
bodysurfing,
bouncing ball (not juggling),
bouncing on bouncy castle,
bouncing on trampoline,
bowling,
braiding hair,
breakdancing,
breaking glass,
breathing fire,
brushing hair,
brushing teeth,
brush painting,
building lego,
bulldozing,
calculating,
calligraphy,
canoeing or kayaking,
capsizing,
card stacking,
casting fishing line,
catching fish,
catching or throwing softball,
changing wheel (not on bike),
cheerleading,
chiseling wood,
chopping wood,
clam digging,
cleaning toilet,
climbing ladder,
climbing tree,
closing door,
coloring in,
combing hair,
contorting,
cooking egg,
cooking on campfire,
cooking sausages (not on barbeque),
counting money,
crossing eyes,
crossing river,
crying,
cumbia,
curling hair,
curling (sport),
cutting apple,
cutting nails,
cutting orange,
dancing ballet,
dancing macarena,
dealing cards,
disc golfing,
dyeing hair,
eating cake,
eating chips,
eating ice cream,
eating nachos,
entering church,
exercising with an exercise ball,
extinguishing fire,
faceplanting,
falling off chair,
feeding birds,
filling cake,
filling eyebrows,
flipping bottle,
flipping pancake,
folding napkins,
gargling,
geocaching,
gold panning,
golf chipping,
golf driving,
golf putting,
gospel singing in church,
grooming dog,
hammer throw,
hand washing clothes,
head stand,
helmet diving,
high fiving,
hitting baseball,
hopscotch,
huddling,
hugging (not baby),
hula hooping,
hurling (sport),
ice climbing,
ice fishing,
jaywalking,
jogging,
juggling balls,
juggling soccer ball,
jumpstyle dancing,
kicking field goal,
kicking soccer ball,
kissing,
kitesurfing,
knitting,
krumping,
laying bricks,
letting go of balloon,
licking,
lifting hat,
lighting candle,
lighting fire,
longboarding,
long jump,
looking at phone,
looking in mirror,
making a sandwich,
making bubbles,
making jewelry,
making latte art,
making snowman,
making tea,
marching,
massaging back,
massaging legs,
massaging person's head,
metal detecting,
milking cow,
milking goat,
mixing colours,
moon walking,
motorcycling,
moving baby,
mowing lawn,
mushroom foraging,
needle felting,
opening coconuts,
opening door,
opening present,
opening refrigerator,
opening wine bottle,
peeling apples,
person collecting garbage,
petting cat,
photocopying,
planting trees,
playing accordion,
playing badminton,
playing bagpipes,
playing beer pong,
playing billiards,
playing blackjack,
playing chess,
playing clarinet,
playing controller,
playing cymbals,
playing drums,
playing field hockey,
playing flute,
playing harp,
playing ice hockey,
playing lute,
playing maracas,
playing marbles,
playing nose flute,
playing oboe,
playing ocarina,
playing piccolo,
playing ping pong,
playing polo,
playing road hockey,
playing rounders,
playing scrabble,
playing shuffleboard,
playing slot machine,
playing squash or racquetball,
playing tennis,
playing ukulele,
playing violin,
playing volleyball,
playing xylophone,
poking bellybutton,
pole vault,
polishing furniture,
polishing metal,
popping balloons,
pouring milk,
pouring wine,
preparing salad,
presenting weather forecast,
pulling espresso shot,
pumping gas,
punching person (boxing),
pushing wheelbarrow,
pushing wheelchair,
push up,
putting on foundation,
putting on lipstick,
putting on mascara,
reading book,
repairing puncture,
riding a bike,
riding mule,
riding or walking with horse,
riding scooter,
riding snow blower,
riding unicycle,
roasting marshmallows,
rolling eyes,
sailing,
scrapbooking,
scrubbing face,
sewing,
shaking hands,
shaking head,
shaping bread dough,
sharpening pencil,
shaving legs,
shearing sheep,
shining flashlight,
shoot dance,
shooting goal (soccer),
shredding paper,
shucking oysters,
shuffling cards,
shuffling feet,
side kick,
silent disco,
situp,
skateboarding,
skiing crosscountry,
slacklining,
slapping,
sled dog racing,
smashing,
smoking hookah,
snorkeling,
spinning plates,
stacking cups,
stacking dice,
steer roping,
stomping grapes,
stretching leg,
surfing crowd,
sweeping floor,
swimming backstroke,
swinging baseball bat,
swinging on something,
sword swallowing,
talking on cell phone,
tap dancing,
tapping guitar,
tapping pen,
tasting wine,
testifying,
throwing snowballs,
throwing tantrum,
tickling,
tie dying,
tobogganing,
tossing coin,
tossing salad,
trimming shrubs,
trimming trees,
twiddling fingers,
tying shoe laces,
unboxing,
using a wrench,
using circular saw,
using puppets,
waiting in line,
walking with crutches,
washing hair,
watching tv,
watering plants,
water skiing,
water sliding,
waving hand,
waxing armpits,
waxing eyebrows,
weaving basket,
windsurfing,
winking,
wood burning (art),
writing,
yawning,
yoga,
zumba.
\end{itemize}

\subsection{Zero-shot Action Localisation}
Here, we initiate two evaluation settings on THUMOS14 and ActivityNet1.3: (A) train on 75\% action categories and test on the remaining 25\% action categories; (B) train on 50\% categories and test on the remaining 50\% categories.

For setting (A) on THUMOS14, the number of training and testing categories is $15$ and $5$, respectively. For setting (B) on THUMOS14, the number of both training and testing action categories is $10$. For setting (A) on ActivityNet1.3, the number of training and testing categories is $150$ and $50$. For setting (B) on ActivityNet1.3, the number of both training and testing categories is $100$.

Under each setting, we conduct $10$ random samplings to split categories for training and testing. Note that, as untrimmed videos in localisation are normally minutes long, splitting datasets based on action categories may incur some situations, where the same video contains both training and testing categories. For this multi-label video, we simply divide it into two videos, one containing only training categories and the other containing only testing categories.

\end{document}